\documentclass[sn-mathphys,Numbered,iicol]{sn-jnl}

\usepackage{graphicx}%
\usepackage{multirow}%
\usepackage{amsmath,amssymb,amsfonts}%
\usepackage{amsthm}%
\usepackage{mathrsfs}%
\usepackage[title]{appendix}%
\usepackage{xcolor}%
\usepackage{textcomp}%
\usepackage{manyfoot}%
\usepackage{booktabs}%
\usepackage{algorithm}%
\usepackage{algorithmicx}%
\usepackage{algpseudocode}%
\usepackage{listings}%

\usepackage{siunitx} 

\def\secref#1{Sec.~\ref{#1}}

\def\figref#1{Fig.~\ref{#1}}
\def\tabref#1{Tab.~\ref{#1}}
\def\eqref#1{Eq.~(\ref{#1})}

\newcommand{\ra}[1]{\renewcommand{\arraystretch}{#1}}

\usepackage{threeparttable}
\usepackage{tablefootnote}

\usepackage{balance}

\usepackage{times}
\usepackage{amsmath,amsfonts}
\usepackage{array}

\usepackage{textcomp}
\usepackage{stfloats}
\usepackage{url}
\usepackage{verbatim}
\usepackage{graphicx}
\usepackage{booktabs} 
\usepackage{bbm} 
\usepackage{mathtools}
\usepackage{pgf} 
\usepackage{etoolbox} 

\usepackage{multicol}

\usepackage{mathtools}
\usepackage{xspace}

\usepackage{multirow}
\usepackage{pbox}

\usepackage[normalem]{ulem}

\definecolor{MK_Two_One}{RGB}{178,24,43} 
\definecolor{MK_Two_Two}{RGB}{239,138,98}
\definecolor{MK_Two_Three}{RGB}{253,219,199}
\definecolor{MK_Two_Four}{RGB}{209,229,240}
\definecolor{MK_Two_Five}{RGB}{103,169,207}
\definecolor{MK_Two_Six}{RGB}{33,102,172} 

\hypersetup{
colorlinks=true
,linkcolor=MK_Two_Six
,citecolor=MK_Two_Six
,filecolor=MK_Two_Six
,urlcolor= MK_Two_Six
,menucolor=MK_Two_Five
,runcolor=MK_Two_Four
,linkbordercolor=MK_Two_One
,citebordercolor=MK_Two_Two
,filebordercolor=MK_Two_Three
,urlbordercolor=MK_Two_Six
,menubordercolor=MK_Two_Five
,runbordercolor=MK_Two_Four
}

\DeclareSIUnit\px{px}

\usepackage{tikz}

\usepackage[abbreviations]{glossaries-extra}

\glssetcategoryattribute{abbreviation}{indexonlyfirst}{true}

\glssetcategoryattribute{abbreviation}{nohyperfirst}{true}

\newabbreviation{auroc}{AUROC}{Area Under the Receiver Operating Characteristic Curve}
\newabbreviation{accuracy}{Acc}{Accuracy}

\newabbreviation{cnn}{CNN}{Convolutional Neural Network}

\newabbreviation{fov}{FoV}{Field of View}
\newabbreviation{fpr}{FPR}{False Positive Ratio}

\newabbreviation{gnn}{GNN}{Graph Neural Network}
\newabbreviation{gcn}{GCN}{Graph Convolutional Network}
\newabbreviation{imu}{IMU}{Inertial Measurement Unit}
\newabbreviation{irl}{IRL}{Inverse Reinforcement Learning}

\newabbreviation{knn}{KNN}{K-Nearest Neighbors}

\newabbreviation{lagr}{LAGR}{Learning Applied to Ground Vehicles}
\newabbreviation{lidar}{LiDAR}{Light Detection and Ranging}

\newabbreviation{mlp}{MLP}{Multi-Layer Perceptron}
\newabbreviation{mpc}{MPC}{Model Predictive Controller}
\newabbreviation{mse}{MSE}{Mean Squared Error}

\newabbreviation{ood}{OOD}{out-of-distribution}

\newabbreviation{rbf}{RBF}{Radial Basis Function}
\newabbreviation{rmp}{RMP}{Riemannian Motion Policies}
\newabbreviation{ros}{ROS}{Robot Operating System}
\newabbreviation{ros1}{ROS~1}{Robot Operating System}
\newabbreviation{roc}{ROC}{Receiver Operating Characteristic}
\newabbreviation{rf}{RF}{Random Forest}

\newabbreviation{sdf}{SDF}{Signed Distance Field}
\newabbreviation{slam}{SLAM}{Simultaneous Localization and Mapping}
\newabbreviation{svm}{SVM}{Support Vector Machine}
\newabbreviation{svc}{SVC}{Support Vector Classifier}
\newabbreviation{wvn}{WVN}{Wild Visual Navigation}

\newabbreviation{vit}{ViT}{Vision Transformer}

\newcommand{\pose}[3]{\mathbf{T}_{\mathtt{#1 #2}}_{#3}}
\newcommand{\rot}[3]{\mathbf{R}_{\mathtt{#1 #2}}_{#3}}
\newcommand{\pos}[2]{{\mathtt{_#1}} \mathbf{p}_{#2}}

\newcommand{\speed}[1]{v_{#1}}
\newcommand{\speedcmd}[1]{\bar{v}_{#1}}

\newcommand{\veldiff}{v_{\text{error}}}

\newcommand{\K}{\mathbf{K}_{3\times3}}

\newcommand{\img}[1]{\mathbf{I}^{#1}}

\newcommand{\feat}[1]{\mathbf{F}^{#1}}

\newcommand{\embed}[1]{\mathbf{f}_{#1}}

\newcommand{\segmask}[1]{\mathbf{M}^{#1}}

\newcommand{\auxsupimage}[1]{\mathbf{S}^{#1}}

\newcommand{\nsupg}{N_{\mathrm{sup}}}
\newcommand{\dsupg}{d_{\mathrm{sup}}}

\newcommand{\dmisg}{d_{\mathrm{mis}}}

\newcommand{\loss}[1]{\mathcal{L}_{\mathrm{#1}}}
\newcommand{\travloss}{\loss{trav}}

\newcommand{\sigmafactor}{k_{\sigma}}
\newcommand{\sigmaanomi}{\sigma_{\mathrm{pos}}}
\newcommand{\muanomi}{\mu_{\mathrm{pos}}}

\newcommand{\SEtwo}{\mathrm{SE(2)}}
\newcommand{\SEthree}{\mathrm{SE(3)}}

\newcommand{\fun}[2]{f_{\mathrm{#1}}\left( #2 \right) }
\newcommand{\sigmoid}[2]{\mathrm{sigmoid}_{\mathrm{#1}}\left( #2 \right) }
\newcommand{\dimss}[2]{#1 \times #2}
\newcommand{\dimsss}[3]{#1 \times #2 \times #3}

\newcommand{\Rn}[1]{\mathbb{R}^{#1}}

\newcommand{\nset}[1]{\mathcal{#1}} 

\robustify{\pose}
\robustify{\rot}
\robustify{\pos}
\robustify{\K}
\robustify{\loss}
\robustify{\feat}
\robustify{\img}
\robustify{\fun}

\newcommand{\trav}[1]{{\tau}_{#1}}

\newcommand{\travthr}[0]{{\tau}_{\text{thr}}}

\newcommand{\dino}[1]{\text{DINO-ViT}{#1}}

\newcommand{\slic}{SLIC}
\newcommand{\stego}{STEGO}

\definecolor{TraversableBlue}{RGB}{49, 54, 149}
\definecolor{UntraversableRed}{RGB}{192, 26, 38}
\definecolor{PaperOrange}{RGB}{251, 151, 39}
\definecolor{PaperMagenta}{RGB}{150, 36, 145}
\definecolor{PaperBlue}{RGB}{67, 110, 176}
\definecolor{PaperCyan}{RGB}{66, 173, 187}

\usepackage{amssymb}
\newcommand{\travsquare}{{\textcolor{TraversableBlue}{$\blacksquare$}}}
\newcommand{\untravsquare}{{\textcolor{UntraversableRed}{$\blacksquare$}}}
\newcommand{\orangesquare}{{\textcolor{PaperOrange}{$\blacksquare$}}}
\newcommand{\magentasquare}{{\textcolor{PaperMagenta}{$\blacksquare$}}}

\newcommand{\cyansquare}{{\textcolor{PaperCyan}{$\blacksquare$}}}

\begin{document}
\title{Wild Visual Navigation: Fast Traversability Learning via Pre-Trained Models and Online Self-Supervision}
\author*[1]{\fnm{Matias} \sur{Mattamala}}\email{matias@robots.ox.ac.uk}
\equalcont{These authors contributed equally to this work.}

\author*[2,3]{\fnm{Jonas} \sur{Frey}}\email{jonfrey@ethz.ch}
\equalcont{These authors contributed equally to this work.}

\author[2]{\fnm{Piotr} \sur{Libera}}
\author[1]{\fnm{Nived} \sur{Chebrolu}}
\author[3,4]{\fnm{Georg} \sur{Martius}}
\author[2]{\fnm{Cesar} \sur{Cadena}}
\author[2]{\fnm{Marco} \sur{Hutter}}
\author[1]{\fnm{Maurice} \sur{Fallon}}

\affil[1]{\orgdiv{Dynamic Robot Systems Group}, \orgname{University of Oxford}, \orgaddress{\street{23 Banbury Road}, \city{Oxford}, \postcode{OX2 6NN}, \state{Oxfordshire}, \country{United Kingdom}}}

\affil[2]{\orgdiv{Robotic Systems Lab}, \orgname{ETH Zurich}, \orgaddress{\street{Leonhardstrasse 21}, \city{Zurich}, \postcode{8092}, \state{Zurich}, \country{Switzerland}}}

\affil[3]{\orgdiv{Autonomous Learning Group}, \orgname{Max Planck Institute for Intelligent Systems}, \orgaddress{\street{Max-Planck-Ring 4}, \city{T\"ubingen}, \postcode{72076}, \state{Baden-W\"urttemberg}, \country{Germany}}}

\affil[4]{\orgdiv{Computer Science Department}, \orgname{University of Tü\"uingen}, \orgaddress{\street{Maria-von-Linden-Strasse 6}, \city{T\"ubingen}, \postcode{72076}, \state{Baden-W\"urttemberg}, \country{Germany}}}

\abstract{
Natural environments such as forests and grasslands are challenging for robotic navigation because of the false perception of rigid obstacles from high grass, twigs, or bushes.
In this work, we present \glsfirst{wvn}, an online self-supervised learning system for visual traversability estimation. The system is able to continuously adapt from a short human demonstration in the field, only using onboard sensing and computing. One of the key ideas to achieve this is the use of high-dimensional features from pre-trained self-supervised models, which implicitly encode semantic information that massively simplifies the learning task. Further, the development of an online scheme for supervision generator enables concurrent training and inference of the learned model in the wild. We demonstrate our approach through diverse real-world deployments in forests, parks, and grasslands. Our system is able to bootstrap the traversable terrain segmentation in less than \SI{5}{\minute} of in-field training time, enabling the robot to navigate in complex, previously unseen outdoor terrains. \\ Code:~\url{https://bit.ly/498b0CV} - Project~page:~\url{https://bit.ly/3M6nMHH}
}

\maketitle

\section{Introduction}
\label{sec:Introduction}
\begin{figure*}
    \centering
	\includegraphics[width=1.0\textwidth]{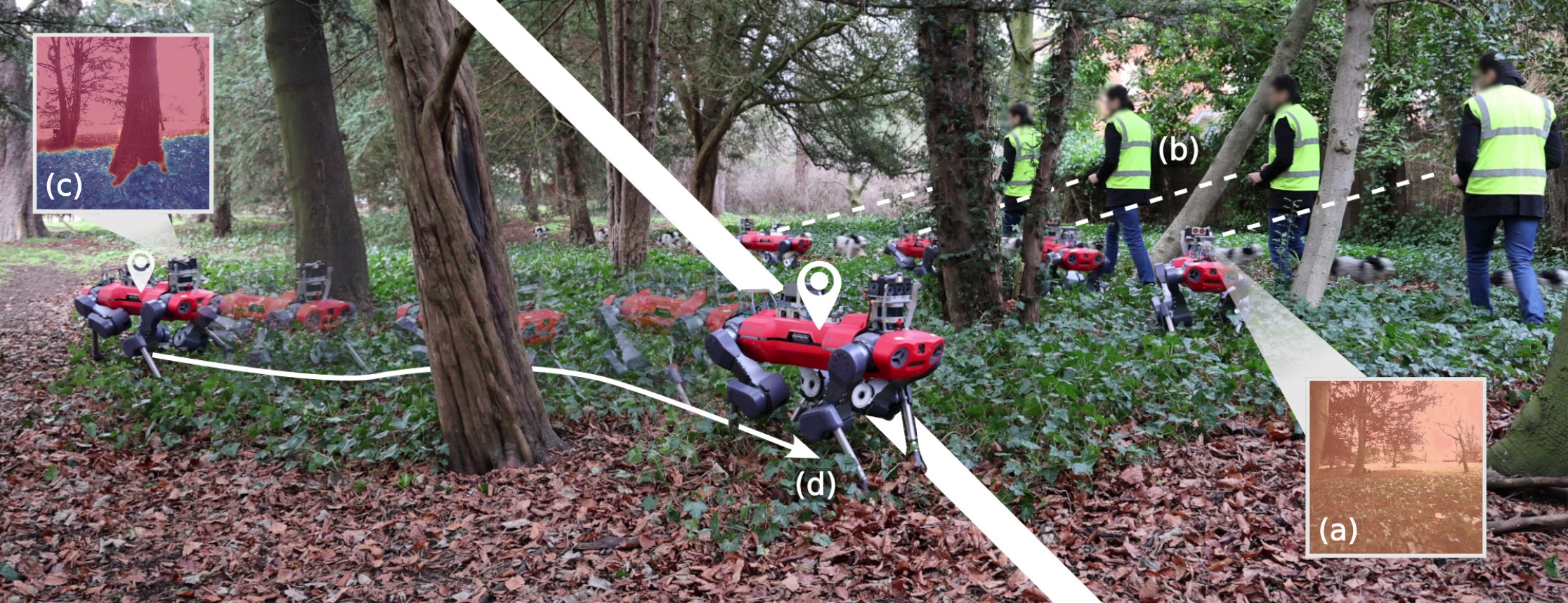}
	\caption{\gls{wvn} learns to predict traversability from images via online self-supervised learning. Starting from a randomly initialized traversability estimation network without prior assumptions about the environment (a), a human operator drives the robot around areas that are traversable for the given platform (b). After a few minutes of operation, \gls{wvn} learns to distinguish between traversable (blue \travsquare{}) and untraversable (red \untravsquare{}) areas (c), enabling the robot to navigate autonomously and safely within the environment (d).}
    \label{fig:header}
\end{figure*}%

Traversability estimation is a core capability needed to allow robots to autonomously navigate in field environments. It is understood as the \emph{affordance}~\citep{Gibson1979} necessary for a robot to navigate within its environment, i.e to understand which areas can be accessed and navigated through and at what cost. While the topic has been widely studied for wheeled or flying robots under the idea of occupancy mapping~\citep{Moravec1985}, the development of new platforms with advanced mobility skills, such as legged robots, prompts a reconsideration of current definitions of traversability, as new and more complex types of natural terrain can be traversed~\citep{Miki2022a}.

Existing approaches, which build upon deep neural models for semantic segmentation \citep{Maturana2017} or anomaly detection \citep{Wellhausen2020}, have demonstrated navigation in off-road environments; however there are recurring problems with the collection and labeling of large amounts of relevant training data.
Self-supervised systems have addressed this challenge by generating labeled datasets from past robot deployments, using classification carried out in hindsight~\citep{Wellhausen2019} or using predictions of the robot motion~\citep{Gasparino2022}. Nevertheless, these previous methods are still trained on robot-specific datasets and subsequently deployed without further adaptation. The \gls{lagr} program~\citep{Kim2006, Hadsell2009} was a first effort towards systems able to adapt in the field, where robots generated its own supervision signals during deployment, facilitating the training of machine learning models within off-road environments.

In this work, we present \gls{wvn}, a system inspired by the aforementioned approaches to achieve self-supervised, online traversability estimation, solely requiring a few minutes of demonstrations in the field. It combines visual input and proprioceptive information to generate supervision signals while the robot operates, enabling it to simultaneously train the traversability model and use it for online inference (\figref{fig:header}). A key idea in \gls{wvn} is exploiting high-dimensional features from pre-trained models. This simplifies the learning task while also exploiting the semantic correspondences implicitly learned by these models via offline self-supervised training on large datasets.

This article extends the system presented by Frey and Mattamala et al.~\cite{FreyMattamala2023}, addressing some of the limitations raised in the original formulation and introducing additional features for system integration and field deployment. The contributions are:
\begin{enumerate}
    \item \textbf{An online, multi-camera self-supervision pipeline} that extends the original approach by enabling the use of multiple vision sources for supervision and inference.
    \item \textbf{The use of pre-trained models} as feature extraction backbones, namely \dino{}\cite{Caron2021} and, in addition to our previous work, \stego{}~\cite{Hamilton2022}. We demonstrate that this eases the overall traversability prediction training process. 
    \item \textbf{A feature sub-sampling strategies} to efficiently process pixel-wise high-dimensional features. We present additional strategies to \slic{}~\cite{Achanta2012} used in our previous work, which further exploit the semantic priors already encoded in the features.
    \item \textbf{Real-world experiments}, demonstrating hardware integration and onboard execution of \gls{wvn} with one and multiple cameras, achieving real-world navigation tasks after minutes of training.
    \item \textbf{Open-source implementation} of the \gls{wvn} system with ROS~\cite{Quigley2009} integration, with a set of baseline model weights trained on diverse environments.
\end{enumerate}

\section{Related Work}
\label{sec:Related work}
\subsection{Traversability from Geometry}
Classical approaches for traversability estimation analyze the geometry of the environment using 3D sensing~\citep{Moravec1985}. Solutions from the DARPA SubT Challenge~\cite{Chung2023}, used representations such as point clouds and meshes to evaluate navigational metrics like risk or stepping difficulty~\citep{Cao2022, Fan2021}.

However, a purely geometric analysis has proven insufficient and data-driven methods have bridged this gap by learning platform-specific traversability from real data or simulations. \citet{Chavez-Garcia2018} used simulations of a ground robot moving on an elevation map. \citet{Yang2021} extended this approach for legged platforms, capturing the risk of failure, energy cost and time required for navigation. Recently, \citet{Frey2022} expanded this approach to volumetric data and massive parallelization in data collection from simulation. Nevertheless, using geometry-only could be insufficient to represent natural growth such as high grass, branches or bushes.

\subsection{Traversability from Semantics}
Semantic segmentation methods aim to address the aforementioned challenges by assigning semantic classes to the representations, with different navigation costs. \citet{Bradley2015} presented a scene understanding system for a legged platform, trained and evaluated using geographically diverse data. \citet{Maturana2017} demonstrated autonomous off-road navigation using semantics projected onto 3D map around a wheeled platform. \citet{Schilling2017} used semantically segmented features that were classified into fixed classes using a random forest classifier. \citet{Belter2019} developed a semantic terrain analysis module to guide a whole-body planner in a multi-legged platform. Recently, \citet{Shaban2022} presented an approach for off-road navigation that learns a dense traversability map from sparse point-clouds, while \citet{Cai2022} mapped terrain semantics to vehicle speed profiles as a proxy for traversability.

Most of these methods rely on pre-trained or fine-tuned semantic segmentation models with pre-defined class labels. In this work we exploit the advances in self-supervised models, such as \dino{}~\citep{Caron2021}, to determine semantically similar regions without manual supervision. 

\subsection{Traversability from Self-supervision}
Self-supervised methods address the challenges of pre-defined classes and costs by using past robot experiences~\citep{Kim2006, Bajracharya2009}. Modern methods rely on deep neural networks trained from weakly supervised data, and the supervision depends on the robot platform. \citet{Wellhausen2019} used the reprojected footholds from a legged robot to provide supervision of walkable areas; \citet{Zurn2021} exploited sounds produced by a wheeled robot moving on different terrain as a proxy for supervision; \citet{Gasparino2022} instead used the receding-horizon trajectory of a \gls{mpc}. 

BADGR~\cite{Kahn2021} predicted future robot states and events from images, including its position and crash probability, which can be interpreted as traversability. TerraPN~\citep{Sathyamoorthy2022} used odometry and IMU signals as supervision to learn a traversability model in $\SI{25}{\min}$ -- including data collection and learning. \citet{GuamanCastro2023} predicted traversability based on IMU supervision conditioned on the velocity of the robot. Recently, \citet{Jung2024} presented a system that shares with \gls{wvn} the use of pre-trained models for self-supervision.

While \gls{wvn} follows similar self-supervision strategies, we aim for concurrent supervision signal generation and learning achieving online adaptation in the field.

\subsection{Traversability from Anomalies}
Anomaly detection methods are motivated by the imbalance of positive and negative samples in self-supervised methods. Instead of training a discriminative model of traversability, they focus on learning generative models of the traversed terrain. This distribution is used as a proxy to set \gls{ood} inputs as untraversable.

\citet{Richter17} trained an autoencoder to predict \gls{ood} scenes from images, switching to safer navigation behaviors when traversing novel environments.
\citet{Wellhausen2020} used multi-modal sensing from haptics, vision and depth to identify anomalies such as flames and water reflections. \citet{Schmid22} show-case the effectiveness of anomaly detection for identifying safe terrain from vision in an off-road driving scenario. Further, \citet{Ji2022} formulated a proactive anomaly detection approach that evaluated candidate trajectories for local planning depending on their probability of failure. 

While we do not explicitly use anomalies to determine traversability, we do use it as a confidence metric to leverage the sparse supervision signals, as it has also been recently explored by \citet{Seo2023}.

\section{Method}
\label{sec:Method}
\subsection{System Overview}
\begin{figure}[t]
    \centering
    \includegraphics[width=\columnwidth]{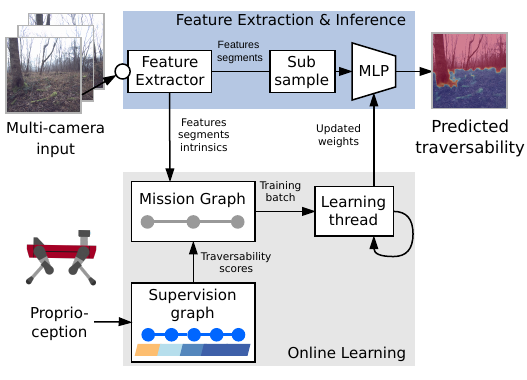}
    \caption{System overview: \gls{wvn} only requires monocular RGB images, odometry, and proprioceptive data as input, which are processed to extract features and supervision signals used for online learning and inference of traversability (see \secref{sec:Method}).}
    \label{fig:block-diagram}
\end{figure}
\begin{table}[t]
    \centering
    \ra{1.1}
    \footnotesize
    \begin{tabular}{c l}\toprule
        Symbol & Definition \\ \midrule
        $\img{}$ & RGB image with height $H$ and width $W$ \\
        $\feat{}$ & Feature map with dim. $\dimsss{E}{H}{W}$, $E=90$ or $384$ \\
        $\segmask{}$ & Weak segmentation mask with height $H$ and width $W$ \\
        $\auxsupimage{}$ &  Reprojected supervision with dim. $\dimss{H}{W}\in [0,1]$ \\
        $\trav{}$ & Traversability score $\in [0,1]$ \\
        $\embed{n}$ & Per-segment embedding of dim. $E=90$ or $384$ \\
        $\trav{n}$ & Per-segment traversability score \\
        \bottomrule
    \end{tabular}
    \caption{Main definitions used in this work}
    \label{tab:notation}
\end{table}
The objective of this work is to design a navigation system that estimates dense traversability scores of the terrain from RGB images. We use a neural network model---\gls{mlp}---trained online, in a self-supervised manner, from supervision signals generated by a robot interacting with its environment. The system should require only a brief demonstration from a human operator for data collection and learning.

\gls{wvn} is implemented as a two-processes system that run at different rates, as shown in \figref{fig:block-diagram}. The \emph{feature extraction \& inference process} processes images from different cameras, extracts visual features, and performs inference to predict the traversability score pixel-wise. The \emph{online learning process} estimates traversability scores from proprioception, generates the supervision signals from hindsight information, and executes an inner training loop to update the traversability prediction model. While the former supplies visual features for training as images are processed, the latter provides the most updated learned model at a fixed time rate.

The main definitions used in the rest of the paper are summarized in \tabref{tab:notation}, and the technical details of each process are provided as follows.

\subsection{Feature Extraction \& Inference}
\label{subsec:feature-extraction}
\begin{figure}[t]
    \centering
    \includegraphics[width=\columnwidth]{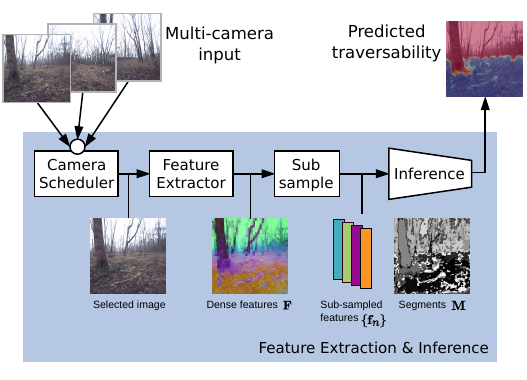}
    \caption{Feature Extraction \& Inference process: The camera scheduler module (\secref{subsubsec:multi-camera}) selects one camera from the available pool, and provides the RGB image to the feature extractor module (\secref{subsubsec:feature-extraction}). This extracts dense visual features $\feat{}$ using pre-trained models. Next, the sub-sample module produces a reduced set of embeddings $\{ \embed{n} \}$ using a subsampling strategy based on a weak segmentation system (\secref{subsubsec:feature-subsampling}). Lastly, the inference module predicts traversability from the image using the embeddings.}
    \label{fig:feature-extraction}
\end{figure}

\subsubsection{Multi-camera processing}
\label{subsubsec:multi-camera}
While the original \gls{wvn} was designed for single-camera processing, it presented limitations during navigation, constraining it to motions within the camera \gls{fov}. 

We enabled multi-camera operation by developing a camera scheduler based on the weighted round-robin algorithm~\cite{Katevenis1991}. This ensured that the system only processes a single camera at a time, depending on priorities specified by the cameras being used for training and inference, or inference-only.

\subsubsection{Feature extraction}
\label{subsubsec:feature-extraction}
After a camera is selected in each cycle of the scheduler, the following steps are camera-agnostic. Given an RGB image $\img{}$, we extract dense, pixel-wise visual feature maps (\emph{embeddings}) $\feat{}$. In contrast to previous works based on fine-tuned \glspl{cnn}, we rely on recent self-supervised network architectures to generate high-dimensional features that encode meaningful semantics without requiring labels.

In our implementation, we integrated the self-supervised \dino{}~\citep{Caron2021}, which provides 384-dimensional pixel-wise feature embeddings. We additionally considered \stego{}~\citep{Hamilton2022}, which uses a \dino{} backbone with additional layers trained with contrastive learning, providing 90-dimensional features and segmentation mask.
Before extracting the features, we resize the input images to a resolution of $224 \times 224$. The resulting dense features $\feat{}$ are too large to be stored in GPU memory for online training. Hence, we introduced feature sub-sampling strategies to reduce the dimensionality of $\feat{}$.

\subsubsection{Feature sub-sampling}
\label{subsubsec:feature-subsampling}

We implemented different sub-sampling strategies to reduce the number of pixel-wise embeddings from $224 \times 224$ to a reduced set of $\sim$100 embeddings $\{ \embed{n} \}$. The strategies use a weak segmentation system to partition the image into a set of segments $\segmask{}$, and then average the embeddings within each segment:

\begin{itemize}
    \item \textbf{SLIC:} In our original implementation and inspired by \citet{Lee2017}, we explored the use of SLIC~\citep{Achanta2012} to reduce the dimensionality to $\sim$100 segments per image. While they are fast to compute, they have the disadvantage of being based on texture only, not necessarily grouping pixels by semantics.
    \item \textbf{\stego{}:} It provides class-free segments, which implicitly encode semantic affinity, which directly defines the segments $\segmask{}$.
    \item \textbf{Random:} We randomly select a set of $100$ embeddings from the feature map. In this case, we have no segments but the feature locations only.
\end{itemize}

\begin{figure}[t]
    \centering
    \includegraphics[width=\columnwidth]{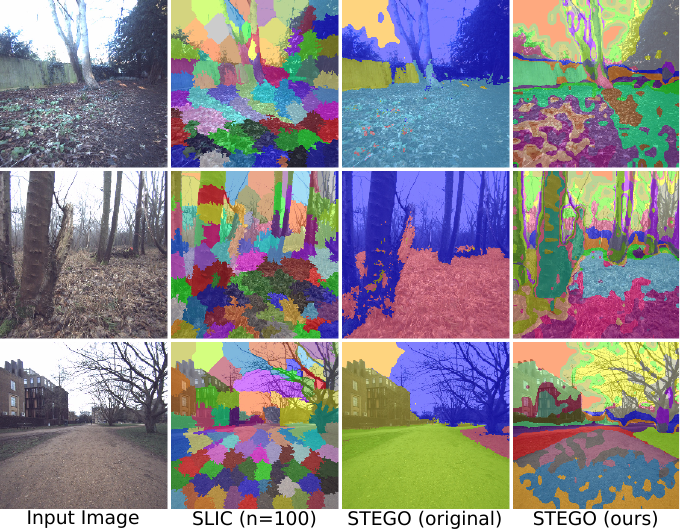}
    \caption{Comparison feature segmentation methods for 3 example images. SLIC over-segments the image, but fails to construct semantically coherent segments (e.g. top row merging fence and ground into a single segment). The \stego{} segmentation aligns with the semantics, but the computation of prototype vectors across a full dataset limits the number of semantic classes, leading to merging of two semantic classes into a single segment (grass and walkway, bottom row). Our modified version of \stego{}, over-segments the image but still provides semantically meaningful segments without pre-setting prototype vectors before deployment.
    }
    \label{fig:feature-subsampling-segmentation}
\end{figure}

The original \stego{} implementation considers the task of semantic segmentation based on a fixed set of classes across a full dataset. To assign each pixel to a semantic class, the authors compute prototype feature vectors across a training dataset offline. 
A pixel can then be assigned to a semantic class based on its cosine-similarity embedding with respect to the identified prototype feature vectors. 
This is not applicable in our scenario, where we do not consider a fixed set of classes, nor can identify suitable prototypes vectors before the mission, given that these would strongly depend on the deployment environment. Instead, we compute a fixed number of prototype features per image using KNN-clustering, which guarantees a fixed number of segments per image.
\figref{fig:feature-subsampling-segmentation} illustrates qualitative examples of the different segments produced by the SLIC and \stego{} methods, and in \secref{subsec:subsampling-strategies}, we experimentally test these different strategies. We open-sourced the full re-implementation of the modified \stego{} version\footnote{GitHub: \url{https://github.com/leggedrobotics/self_supervised_segmentation}}.

After this sub-sampling step, the subset of embeddings $\{ \embed{n} \}$ and their image locations, segments $\segmask{}$ and camera intrinsics are shared with the \emph{online learning} process for training (\secref{subsec:online-learning}).

\subsubsection{Inference}
Lastly, this process provides predictions of traversability for all the incoming images from the different cameras. This is achieved by inferencing the \gls{mlp} model, which is updated at a fixed rate by the online learning process (\secref{subsec:platform-description}). The model predicts traversability from the embeddings $\{ \embed{n} \}$ using two different approaches:
\begin{itemize}
\item \textbf{Segment-wise inference:} This is the approach implemented originally~\cite{FreyMattamala2023}, which predicts a traversability score $\trav{n}$ for each embedding $\embed{n} $, and assigns the same score for all the pixels corresponding to the given segment.
\item \textbf{Pixel-wise inference:} Alternatively, we predict fine-grained traversability from the dense features $\feat{}$, given that the \gls{mlp} forward pass can be executed with low-latency for a batch of features.
\end{itemize}
\secref{subsec:subsampling-strategies} provides qualitative examples of the improvements that each method provides when the system is deployed in different natural environments.

\subsection{Online Learning}
\label{subsec:online-learning}

\subsubsection{Traversability Score Generation}
\label{subsubsec:traversability-score-generation}
Defining which terrain is traversable or not depends on the capabilities of the specific platform. We define a continuous \emph{traversability score} $\trav{} \in [0,1]$, where $0$ is untraversable and $1$ is fully traversable. We use the terrain \emph{traction}~\cite{Cai2023}, which measures the discrepancy between the robot's current linear velocity as estimated by the robot $(v_x, v_y)$, and the reference velocity command $(\bar{v}_x, \bar{v}_y)$ given by an external human operator or planning system.

We define the mean squared velocity error as:
\begin{equation}
    \label{eq:velocity-error}
    \veldiff{} = \frac{1}{2} \left( \left( \speedcmd{x} -  \speed{x}  \right)^{2} + \left( \speedcmd{y} -  \speed{y}  \right)^{2} \right) \in \Rn{}
\end{equation}
We smooth $\veldiff{}$ with a 1-D Kalman Filter before passing it through a sigmoid function to obtain a valid traversability score:
\begin{equation}
    \label{eq:traversability}
    \trav{} = \sigmoid{}{ - k\left( \veldiff{} - v_{\text{thr}} \right) }
\end{equation}
with $k$ the steepness of the sigmoid, and $v_{\text{thr}}$ the midpoint of the sigmoid that assigns a traversability score of $0.5$. These values are calibrated depending on the motion specifications of each platform and determine how the velocity error is stretched to the $[0,1]$ interval.

\subsection{Supervision and Mission Graphs}
\label{subsec:sup-mission-graphs}

\begin{figure*}[t]
    \centering
    \includegraphics[width=\textwidth]{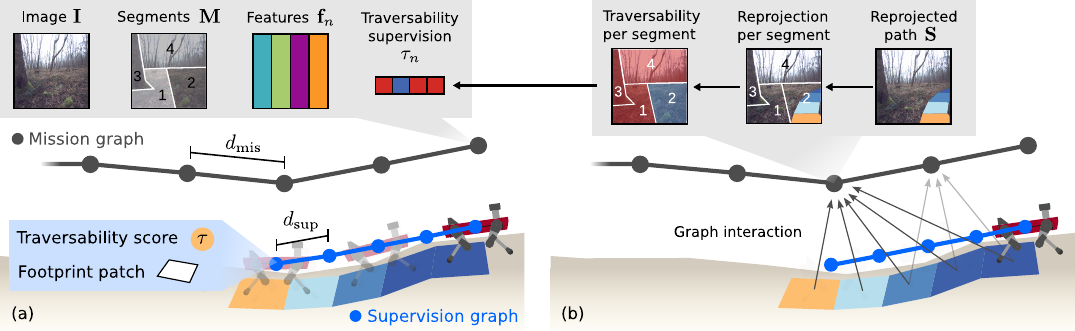}
    \caption{Supervision and mission graphs: (a) Information stored in each graph over the mission. While the Supervision Graph only stores temporary information about the robot's footprint in a sliding window, the Mission Graph saves the data required for online learning over the full mission. The color of the footprint patches indicates the generated traversability score. 
    (b) The interaction between graphs updates the traversability in the mission nodes by reprojecting the robot's footprint and traversability scores.}
    \label{fig:supervision-and-mission-graphs}
\end{figure*}

The system generates supervision signals by accumulating information in hindsight, during operation. Our approach is inspired by graph-based SLAM pipelines that leverage both local and global graphs to integrate measurements: we maintain a \emph{Supervision Graph} to store short-horizon traversability data, and a global \emph{Mission Graph} which stores the generated training data during a mission, shown in \figref{fig:supervision-and-mission-graphs}.

\subsubsection{Supervision Graph}
The supervision graph stores within its nodes information about the current time, robot pose, and estimated traversability score (\secref{subsubsec:traversability-score-generation}). This graph is implemented as a ring buffer, which only keeps a fixed number of nodes $\nsupg{}$, separated from each other by a distance $\dsupg{}$.

The stored information is a footprint track with traversability scores $\trav{}$. It is used to associate traversability scores with features by projecting the footprint track into the previous camera viewpoints.

\subsubsection{Mission Graph}
The mission graph stores all the information required for online training. The mission nodes are added to the graph after feature extraction if the distance with respect to the last added node is larger than $\dmisg{}$. Each mission node contains the RGB image $\img{}$, the weak segmentation mask $\segmask{}$ and per-segment features $\embed{n}$ with their corresponding traversability supervision $\trav{n}$.

\subsubsection{Supervision generation}
When a new mission node is added, we update the supervision labels $\trav{n}$ by reprojecting the footprint track and corresponding traversability scores $\trav{}$ onto all the images of the mission nodes within a fixed range (\figref{fig:supervision-and-mission-graphs}b).

Each mission node then has an auxiliary image with the reprojected path, $\auxsupimage{}$. We use the weak segmentation mask $\segmask{}$ to assign per-segment traversability supervision values $\trav{n}$ by averaging the score over each segment. Segments that do not overlap with the reprojected footprint track are set to zero (i.e untraversable). The outcome are pairs of per-segment features $\embed{n}$ and traversability score $\trav{n}$ for each mission node, used for training.
%
\subsection{Traversability and Anomaly Learning}
\label{subsec:trav-learning}

We train a small neural network in an online fashion that determines the feature traversability score $\trav{n}$ from a given segment feature $\embed{n}$. This reduces the visual traversability estimation problem to simple regression task. Further, we model the uncertainty about the unvisited (and hence, unlabeled) areas by using anomaly detection techniques to bootstrap a confidence estimate.

First, we elaborate on how a confidence score for a feature is obtained, then we describe the traversability estimation learning task. 
\subsubsection{Confidence Estimation}
\label{subsubsec:confidence-estimation}
To obtain a segment-wise confidence estimate, we aim to learn the distribution over all traversed segment features $\embed{n}$. An encoder-decoder network $f_{\text{reco}}^{\theta_r}$ is trained to compress the segment feature $\embed{n}$ into a low dimensional latent space and reconstruct the original input features $\embed{n}$. 
The reconstruction loss is given by the \gls{mse} between the predicted features and the original feature compute over all channels $E$:
\begin{equation}
\mathcal{L}_{\text{reco}}(\embed{n}) =  \begin{cases} \frac{1}{E} \displaystyle\sum_{e}{ \lVert f_{\text{reco}}^{\theta_r}(\embed{n,e}) - \embed{n,e}  \rVert^{2}} & \text{if traversed,} \\
0 & \text{otherwise}
\end{cases}
\end{equation}

This ensures that the network only learns to reconstruct the embeddings that are labeled, in an anomaly detection fashion. Consequently, the trained network reconstructs known (\emph{positive}) feature embeddings, i.e. similar to the traversable segments, with small reconstruction loss; feature embeddings of unknown (\emph{anomalous}) segments the network was never tasked to reconstruct, such as trees or sky, induce a high reconstruction loss. 

The unbounded reconstruction loss $\mathcal{L}_{\text{reco}}$ for a segment is mapped to a confidence measure $c(\mathcal{L}_{\text{reco}}) \in [0,1]$ by first identifying the mode of the traversed segment losses. For this we fit a Gaussian distribution $\mathcal{N}(\muanomi{}, \sigmaanomi{})$ over the reconstruction losses per batch of the traversed segments (i.e, positive samples):
\begin{align}
\muanomi{} &= \frac{1}{n_{\text{trav}}} \sum_{n \in\, \nset{T}} \mathcal{L}_{\text{reco}}(\embed{n}), \\
\sigmaanomi{} &= \sqrt{ \frac{1}{n_{\text{trav}}} \sum_{n \in \nset{T}}  \left( \mathcal{L}_{\text{reco}}(\embed{n})-\muanomi{} \right)^2  }
\end{align}
with $\nset{T}$ being the set of segments that were traversed, i.e. have a valid traversability score $\trav{n}$ computed from robot sensing data, and  $n_{\text{trav}}$ is the total number of traversed segments. We set the segment confidence to $1$ if the loss of the segment is smaller than $\muanomi{}$ and otherwise we set it by evaluating the unnormalized Gaussian likelihood:
\begin{align}
c(\mathcal{L}_{\text{reco}}(\embed{n})) = & \exp{\left( \frac{ (\mathcal{L}_{\text{reco}}(\embed{n}) -\muanomi{})^2}{2(\sigmaanomi{}\,\sigmafactor{})^2} \right) },
\end{align}
where we introduce the tuning parameter $\sigmafactor{}$, which allows to scale the confidence.

\subsubsection{Traversability Estimation}
We train a small network $f_{\text{trav}}^{\theta_t}$ with a single channel output to regress on the provided segment traversability score $\trav{}$.
For the untraversed segments with unknown traversability score, we follow a conservative approach setting $\trav{}=0$ but using the confidence score to scale their overall contribution. 
The loss for traversability estimation is computed using the confidence-weighted \gls{mse}:

\begin{align}
  \displaystyle
  \travloss{}(\embed{}) = &  
    \underbrace{\sum_{n \in\, \nset{T}} \lVert f_{\text{trav}}^{\theta_t}(\embed{n}) - \trav{n} \rVert^{2}}_{\text{\parbox{5cm}{\centering Contribution of traversed (\emph{labeled}) segments}}} \\
    & +
    \underbrace{\sum_{n \in\, \nset{T}^{C}} (1-c(\embed{n}))\, \lVert f_{\text{trav}}^{\theta_t}(\embed{n}) - 0 \rVert^{2}}_{\text{\parbox{5cm}{\centering Contribution of untraversed segments}}},
\end{align}
with $\nset{T}$ the set of traversed segments; $\nset{T}^{C}$ is the complement set of untraversed segments. This formulation enables the learning process to ``overwrite" previously unknown samples as new data is used for training:
\begin{itemize}
  \item If the segment $n$ was traversed: it will contribute to the loss using the assigned traversability score: $\travloss{}(\embed{n}) = \lVert f_{\text{trav}}^{\theta_t}(\embed{n}) - \trav{n} \rVert^{2}$
  \item If the segment $n$ was untraversed and it does not resemble a positive sample: its confidence will be low $c(\embed{n}) \rightarrow 0$ and $\travloss{}(\embed{n}) \rightarrow \lVert f_{\text{trav}}^{\theta_t}(\embed{n}) - 0 \rVert^{2}$
  \item If the segment $n$ was untraversed but it does resemble a positive sample: its confidence $c(\embed{n}) \rightarrow 1$ and $\travloss{}(\embed{n}) \rightarrow 0$, effectively not contributing to the loss anymore. This motivates the network to learn the traversability score measured by physically interacting with the segment as a opposed to being too pessimistic.
\end{itemize}

As we aim to provide the estimated traversability as input for a local planning system, we automatically define a threshold to determine the traversable and untraversable areas. We select a traversability threshold $\travthr{}$ by measuring the current performance of the system in a self-supervised manner. We compute the \gls{roc} throughout training by classifying all segments with confidence under $0.5$ as negative and traversed segments as positive labels. Then, we decide on the traversability threshold only by setting the desired \gls{fpr}.

\subsubsection{Implementation details}
We implemented $f_{\text{reco}}^{\theta_r}$ and $f_{\text{trav}}^{\theta_t}$ as a two-layer \glspl{mlp} with [256, 32] unit dense layers and ReLU non-linear activation functions.
Both networks share the weights of the hidden layers. 
$f_{\text{reco}}^{\theta_r}$ has a reconstruction head with $E$ output neurons and $f_{\text{trav}}^{\theta_t}$ a single channel traversability head followed by a sigmoid activation. 
The 32-channel hidden layer functions as the bottleneck of the encoder-decoder structure.
The total loss per segment during training is given by:
\begin{equation}
\mathcal{L}_{\text{total}}(\embed{}) = w_{\text{trav}}\mathcal{L}_{\text{trav}}(\embed{})  + w_{\text{reco}}\, \mathcal{L}_{\text{reco}}(\embed{}) .
\end{equation}
with $w_{\text{trav}}$ and $w_{\text{reco}}$ allowing to weigh the traversability and reconstruction loss respectively. We used Adam~\citep{Kingma2015} to jointly train the networks with a fixed constant learning rate of $0.001$. For a single update step, 8 valid mission nodes are randomly chosen to form a data batch, where we defined a node as valid if at least a single segment of the node has non-zero traversability score. For all our experiments we set $\sigmafactor{}=2$, $w_{\text{trav}}=0.03$, $w_{\text{reco}}=0.5$ and use a maximum \gls{fpr} of $0.15$ to determine the traversability threshold. Please refer to our previous publication~\citep{FreyMattamala2023} for ablation studies of the different parameter and design choices.

\section{Closed-loop Integration}
\label{sec:closed-loop_integration}
We integrated the learned traversability estimate into a standard navigation pipeline to achieve autonomous navigation with a quadrupedal platform. The details are explained as follows.

\subsection{Local terrain mapping}
\label{subsec:local-terrain-mapping}
In order to use the predicted traversability for navigation tasks, we used an open-source terrain mapping framework~\citep{Miki2022b, Erni2023} that produced a robot-centric 2.5D elevation map from the onboard depth cameras and LiDAR sensing. The framework enabled the fusion of the predicted traversability images via raycasting, taking into account the occlusions with the terrain, as well as temporal fusion of the traversability information via exponential averaging.

\subsection{Local planning}
\label{subsec:local-planning}
We used the projected visual traversability as a cost map for a reactive local planner~\citep{Mattamala2022} to generate a $\SEtwo{}$ twist command to drive the robot towards a goal while avoiding untraversable terrain. The twist command was the input to a robust learning-based locomotion controller~\citep{Miki2022a}, which is able to traverse rough terrain typically inaccessible to wheeled robots.

\subsection{Autonomous Path Following}
\label{subsec:carrot}
Lastly, we implemented a navigation strategy to guide the robot in path-like environments. The method continuously spawned new goals in front of the robot by finding the furthest traversable position in the local terrain map, within the \gls{fov} of the front-facing camera. This strategy was used to motivate simple autonomous navigation and exploration without requiring a global planner and a large-scale representation of the environment.

\section{Experiments}
\label{sec:Experiments}
\subsection{Platform Description}
\label{subsec:platform-description}
For our experiments we used the ANYbotics ANYmal~C and D legged robots. In both configurations the robots were equipped with an additional NVidia Jetson Orin AGX. We used the manufacturer's state estimator to obtain $\SEthree{}$ pose and body velocity measurements. The LiDAR and depth cameras available on the robots were only used for the local terrain mapping module (\secref{subsec:local-terrain-mapping}).

For the ANYmal~C experiments, we used a single global shutter, wide \gls{fov} camera from the Sevensense Alphasense Core unit. For the ANYmal~D experiments, we used the RGB images from the integrated front and rear wide-angle cameras.

\gls{wvn} was implemented in pure Python code using PyTorch~\cite{Paszke2019} and ROS~1 \cite{Quigley2009}. Both processes ran on the Jetson Orin, and were implemented as separate ROS nodes. Inter-process communication was implemented using ROS publisher-subscriber paradigm, while the trained model weights were shared via write-read operations every \SI{5}{\second} for simplicity.

\subsection{Real-world deployments}
We executed different deployments to validate \gls{wvn} in real environments in terms of adaptation to new scenes, the advantages of the visual traversability estimation compared to purely geometric, and autonomous navigation demonstrations.
%
\vspace{0.3em}
\subsubsection{Fast Adaptation on Hardware}
\label{subsubsec:loop-adaptation}
Our first experiment involved teleoperating the ANYmal~C robot around 3 loops in University Parks, Oxford, UK, to evaluate the fast adaptation capabilities of \gls{wvn} when walking on grass and dirt, on open areas, and around trees.

\begin{figure}[t]
    \centering
    \includegraphics[width=\columnwidth]{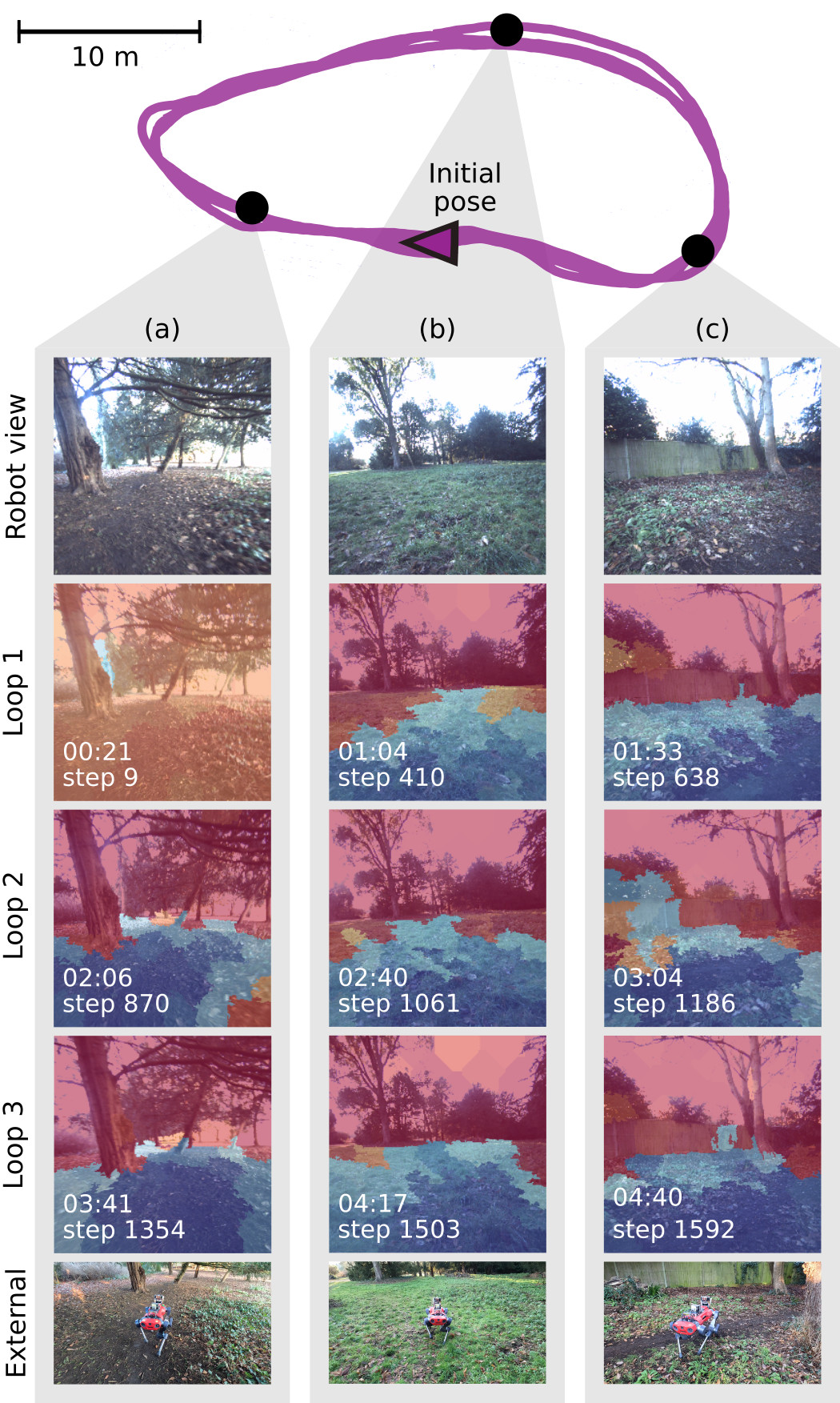}
    \caption{Adaptation on real hardware: We tested the online adaptation capabilities of our system by teleoperating the robot to complete 3 loops in a park (top, route shown in \magentasquare{}). The columns show different parts of the loop (a,b,c); each row displays the improvement of the traversability estimate over time and training steps.}
    \label{fig:loop-adaptation}
\end{figure}

\figref{fig:loop-adaptation} illustrates the main outcomes of the experiment, showing that the system learned to predict robot-specific traversability over the 3 loops while running onboard. 
Section \textbf{(a)} shows how the robot starts with a very poor segmentation after 9 steps of training (\SI{21}{\second}) but this greatly improves after 800 steps (\SI{2}{\minute}), where it can correctly segment the dirt as traversable terrain while keeping the tree untraversable. Similar behavior occurs in section \textbf{(b)} in which the segmentation is conservative at the beginning  but it extends across the other grass patches in later iterations. Section \textbf{(c)} also illustrates some issues related to the SLIC segmentation, as some segments of the wooden wall (step 1186) are incorrectly clustered with patches of the grass, which is not observed in the other captures.

\vspace{0.3em}
\subsubsection{Benefits of Visual Traversability vs Geometric Methods}
\label{subsubsec:traversability-comparison}
\begin{figure*}[t]
    \centering
    \includegraphics[width=\textwidth]{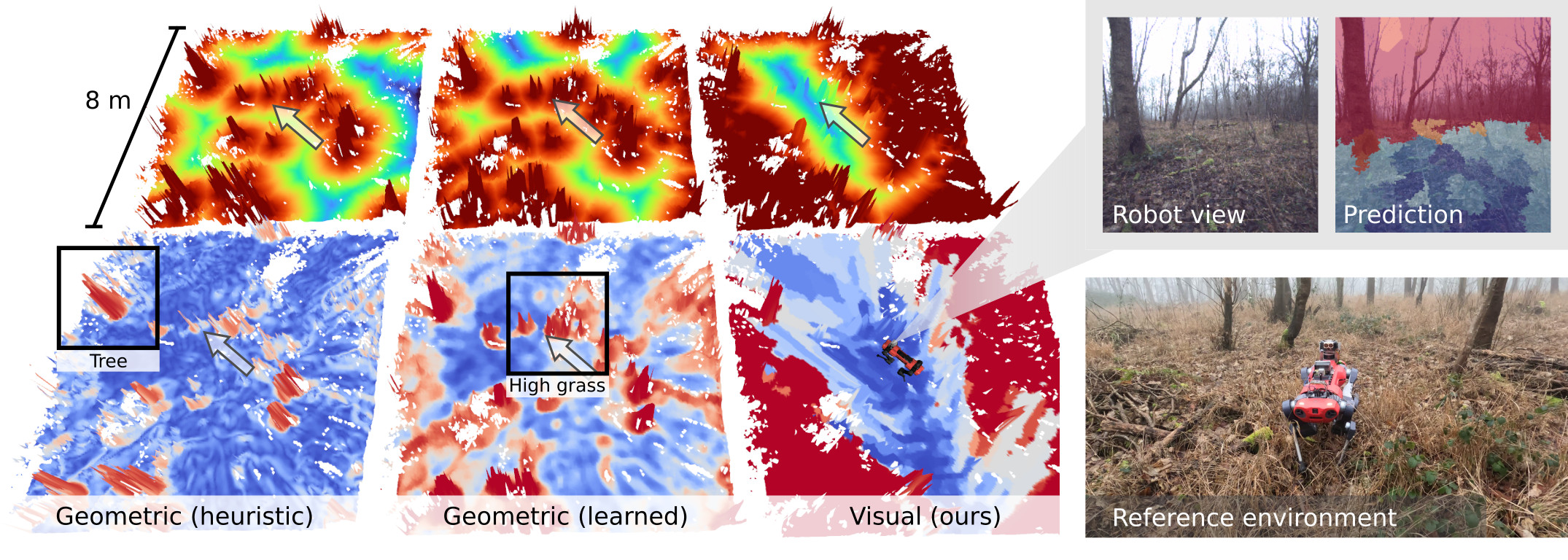}
    \caption{Visual vs geometric traversability: Illustration of traversability map (bottom row) and corresponding \gls{sdf} (top row) for three different traversability estimation methods applied to the same terrain patch. Our visual traversability estimate provides clear advantages for local planning compared to geometric methods, where the latter get heavily affected by traversable high grass or branches (bottom row). This is evident when comparing the \gls{sdf}{'s}, where geometry-based methods are more sensitive to the spikes produced by high grass areas (top row).}
    \label{fig:traversability-comparison}
\end{figure*}
Our second experiment aimed to illustrate the advantages of visual traversability estimation in challenging natural environments. We teleoperated the ANYmal~C robot around high grass, loose branches, and bushes in Wytham Woods, Oxford, UK. \figref{fig:traversability-comparison}, bottom right, shows a representative shot of the experiment, the forward-facing camera image and \gls{wvn}{'s} prediction.

To compare the different traversability methods, we used the terrain mapping module (\secref{subsec:local-terrain-mapping}), as it allowed us to compare geometry-only and visual traversability. We compared against two geometric methods that are real-time capable and have been used in previous works:
\begin{itemize}
\item Geometric method based on heuristics such as height and slope of the terrain~\cite{Wermelinger2016}.
\item Geometric method based on a learned model of traversability, which is part of the terrain mapping system~\cite{Miki2022b}.
\item Visual traversability provided by \gls{wvn}, raycasted onto the terrain map.
\end{itemize}
The geometric methods only require an elevation representation of the surface to determine traversability from the 2.5D geometry. For \gls{wvn} we executed a training procedure by driving the robot around the environment for a few minutes, only using images from the forward-facing camera.

\figref{fig:traversability-comparison} illustrates the output \emph{traversability map} obtained by all the methods (bottom), as well as the corresponding \gls{sdf}{s} generated from them (top). The geometric methods correctly determine the trees as untraversable areas. Our system is also able to successfully discriminate the trees, confirming the findings observed in \secref{subsubsec:loop-adaptation}. Furthermore, the advantages of \gls{wvn} are observed in high-grass areas, which are represented as elevation spikes in the map that are classified as untraversable by the geometric methods but not by our visual traversability estimate. These differences become more evident in the \gls{sdf}{s} where all the areas with low traversability scores become obstacles.
%
\vspace{0.3em}
\subsubsection{Point-to-point Autonomous Navigation Between Trees}
\begin{figure*}[t]
    \centering
    \includegraphics[width=\textwidth]{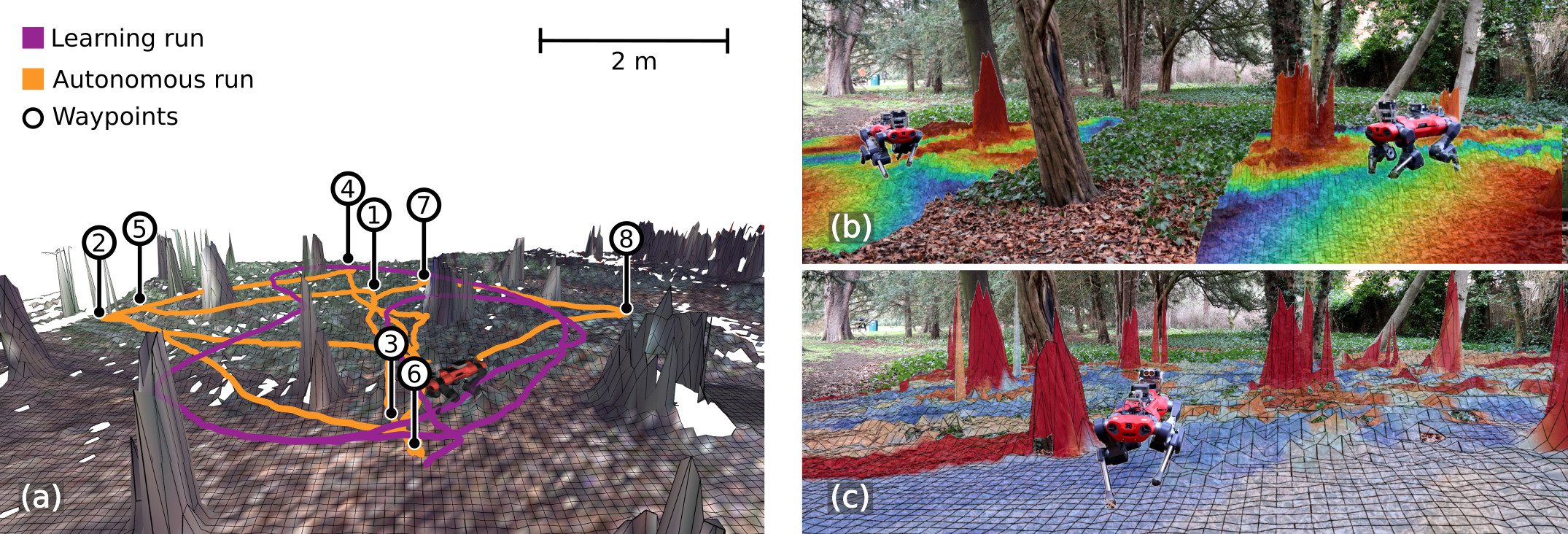}
    \caption{Point-to-point autonomous navigation: (a) After teleoperating the robot for \SI{2}{\minute} (path shown in \magentasquare{}), we successfully achieved autonomous navigation in a woodland environment (path shown in \orangesquare{}). (b) Some of the \gls{sdf}{s} generated from the predicted traversability during autonomous operation. (c) Global 2.5D reconstruction of the testing area and predicted traversability, generated in post-processing to illustrate the capabilities of our approach.}
    \label{fig:real-world-point-to-point}
\end{figure*}
We executed closed-loop navigation tasks to demonstrate that \gls{wvn} can easily adapt to a new environment, and the learned traversability estimate can be used to deploy the robot autonomously. 

We taught the ANYmal~C robot to navigate in a woodland area containing dirt, high grass, and trees. A human operator drove the robot for \SI{2}{\minute} through loose dirt and grass---an area that can be easily traversed by the legged platform. Then we commanded the local planner to execute autonomous point-to-point navigation avoiding obstacles, only using the visual traversability for closed-loop planning \secref{sec:closed-loop_integration}.

\figref{fig:real-world-point-to-point} illustrates the scene used for the experiment and the trajectories used for training and testing autonomous navigation. The robot successfully managed to reach 8 out of 8 goals, where the human operator deliberately chose targets behind trees to challenge the system. This was achieved even though neither geometry nor any additional assumptions about the environment were used during training.

We also show some examples of the \gls{sdf}{s} generated during operation used by the local planner in \textbf{(b)}, which indicate the trees as obstacles. Lastly, in post-processing we fused the predicted traversability measures into a complete map in \textbf{(c)}, which correctly aligned with the trees. However, given that in this experiment we used the SLIC segmentation method from our previous work, we observed some obstacle artifacts. This limitation is addressed in \secref{subsec:mulit}, where we deploy our multiple-camera setup and the novel segmentation and pixel-wise prediction method.

%
%
\vspace{0.3em}
\subsubsection{Kilometer-scale Autonomous Navigation in the Park}
\label{subsubsec:exp-real-world-path-following}
\begin{figure*}[t]
    \centering
    \includegraphics[width=1.0\textwidth]{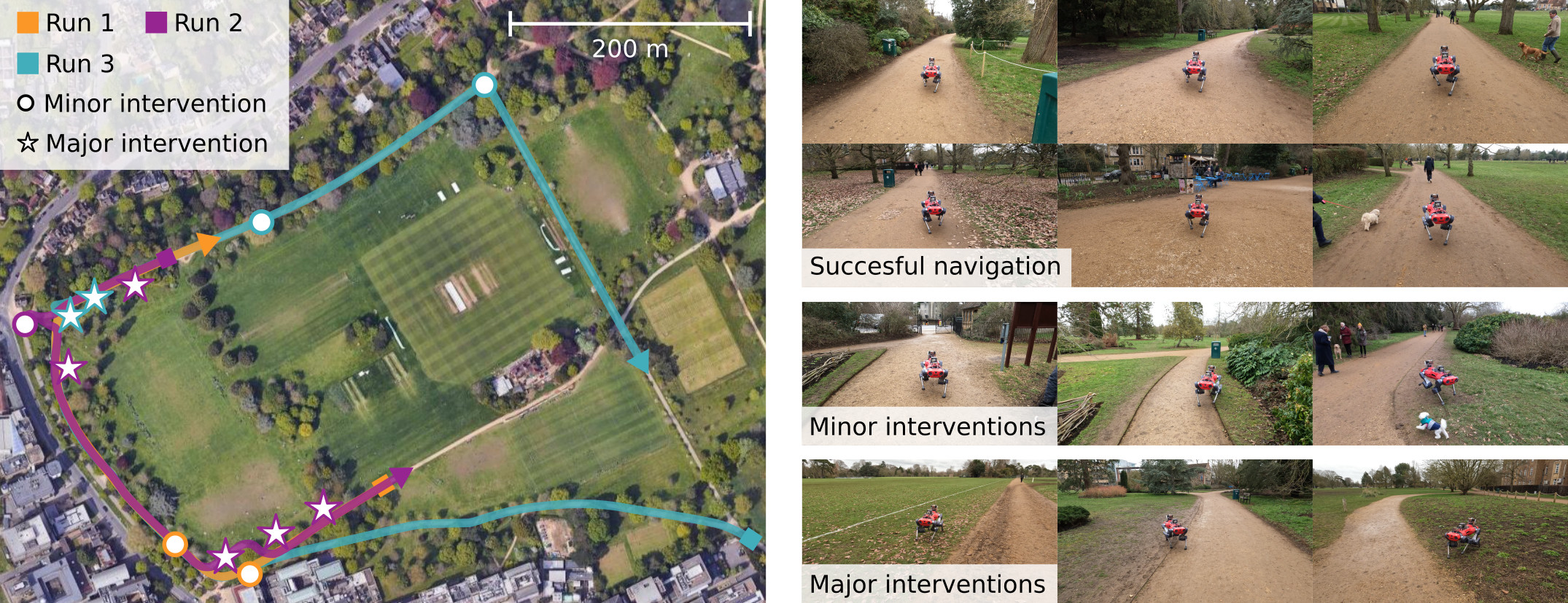}
    \caption{Kilometer-scale navigation: We deployed our system to learn to segment the footpath of a park after training for a few steps. We executed 3 runs starting from different points in the park: \orangesquare{} \emph{run 1} (\SI{0.55}{\kilo\meter}), \magentasquare{} \emph{run 2} (\SI{0.5}{\kilo\meter}), and \cyansquare{} \emph{run 3} (\SI{1.4}{\kilo\meter}). Minor interventions were applied to guide the robot in intersections; major interventions ($\mathbf{\star}$) were required for some areas when the robot miss-classified muddy patches for the path.
    }
    \label{fig:real-world-path-following}
\end{figure*}
We demonstrated that \gls{wvn} enabled preference-aware path-following behavior as a result of the human demonstrations and the online learning capabilities of the system. This experiment was also executed with the ANYmal~C platform.

We executed 3 runs in a footpath at University Parks, Oxford, UK. Similarly to our previous experiments, we trained the system for less than \SI{2}{\minute} along the footpath. We then disabled the learning process to ensure that the predicted traversability strictly mimics the human preference during the demonstration run. The autonomous path following system from  \secref{subsec:carrot} was used to guide the robot forward along the path. 

In the 3 runs the robot was able to follow the path for hundreds of meters---mostly staying in the center of the path, avoiding grass, bushes, benches, and pedestrians. \figref{fig:real-world-path-following} shows the trajectories followed in each run, starting from different points in the footpath. For runs $1$ and $3$ we used the same parameters, $\sigmafactor{}=2$ and \gls{fpr}$=0.15$. In run 2 we relaxed the parameters to $\sigmafactor{}=3$ and \gls{fpr}$=0.3$, producing a less conservative behavior that drove the robot to other visually similar areas in the park (mud patches) requiring manual intervention to correct the heading. When the robot approached an intersection we adjusted, if necessary, the heading to follow the desired footpath.

Overall, we achieved autonomous behavior that would have been difficult to achieve using only geometry, as the path boundaries were often geometrically indistinguishable
. On the other hand, instead of training and using a semantic segmentation system to learn \emph{all} the possible traversable classes in the park (pavement, gravel path, roadway or grass), we showed that this short teleoperated demonstration of the gravel footpath was sufficient for \gls{wvn} to generate semantic cues to achieve the desired path following behavior.

\subsubsection{Multi-camera Deployment from Indoor to Outdoor Environments}
\label{subsec:mulit}
\begin{figure*}[t]
    \centering
    \includegraphics[width=0.8\textwidth]{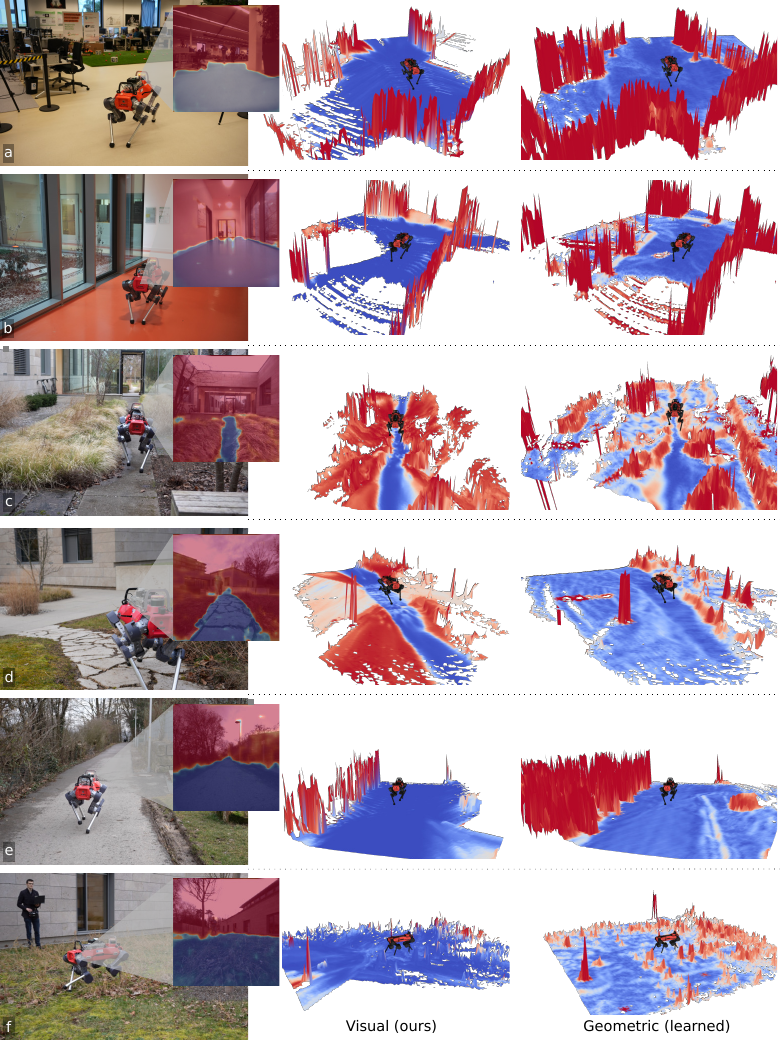}
    \caption{Deployment with multi-camera setup. Left: Real scene and visual traversability prediction. Center: Visual traversability projected on the local terrain map. Right: Geometric traversability computed from elevation map. The robot was teleoperated throughout the experiment. Each example (a) - (f) is sequentiall and it is discussed in detail in \secref{subsec:mulit}.}
    \label{fig:mpi_deployment}
\end{figure*}

This last experiment demonstrates the adaptation capabilities of \gls{wvn} and the new multi-camera integration on the ANYmal D quadruped. The deployment was executed at the Max Planck Institute in Tübingen, Germany. We deployed \gls{wvn} using \stego{} segmentation and features during training and perform the inference pixel-wise. Throughout the \SI{7}{min} teleoperated session, we provide snapshots of the environment, the traversability predictions, as well as the visual and geometric traversability, illustrated in \figref{fig:mpi_deployment}.

The deployment started within a laboratory setting. Upon covering $\SI{\sim 8}{m}$ \textbf{(a)}, \gls{wvn} correctly identified the floor as traversable. Transitioning to a corridor after $\SI{\sim 40}{m}$ \textbf{(b)}, the visual traversability accurately classified windows and closed glass doors as impassable, which the geometric traversability erroneously report as traversable. 
When exiting the building, \gls{wvn} correctly predicted the paved walkways as traversable, which can be seen within the courtyard \textbf{(c)} at $\SI{\sim 86}{m}$, outdoor walkway \textbf{(d)} at $\SI{\sim 127}{m}$, and the paved road \textbf{(e)} at $\SI{\sim 148}{m}$. When entering a small grass area with sparse vegetation after walking for a few minutes, ours correctly identified the field as traversable, while the geometric traversability fails to distinguish between trees and penetrable vegetation \textbf{(f)} at $\SI{\sim 260}{m}$.

The integration of traversability estimates from both cameras enabled us to update the traversability to the front and the back of the robot. This allowed to overcome the restricted field of view limitations shown in \figref{fig:traversability-comparison}. Multiple cameras also allow for more reactive behavior in dynamic environments, where it is crucial for planning to update the belief about the environment constantly.

\subsection{Offline analysis}
We executed two offline experiments to assess the differences of the feature sub-sampling and inference methods. These experiments were executed in post-processing using logs of previous real-world experiments, on an Nvidia Quadro T2000 Laptop GPU with Intel i7-10875H CPU.

\subsubsection{Segments vs Pixel-wise Inference}
First, we compared the visual traversability prediction differences when performing segmen-wise and pixelwise inference. We ran \gls{wvn} in post-processing, on the recorded logs from the \secref{subsubsec:traversability-comparison} and \secref{subsubsec:loop-adaptation} experiments.

\begin{figure*}[t]
    \centering
    \includegraphics[width=\textwidth]{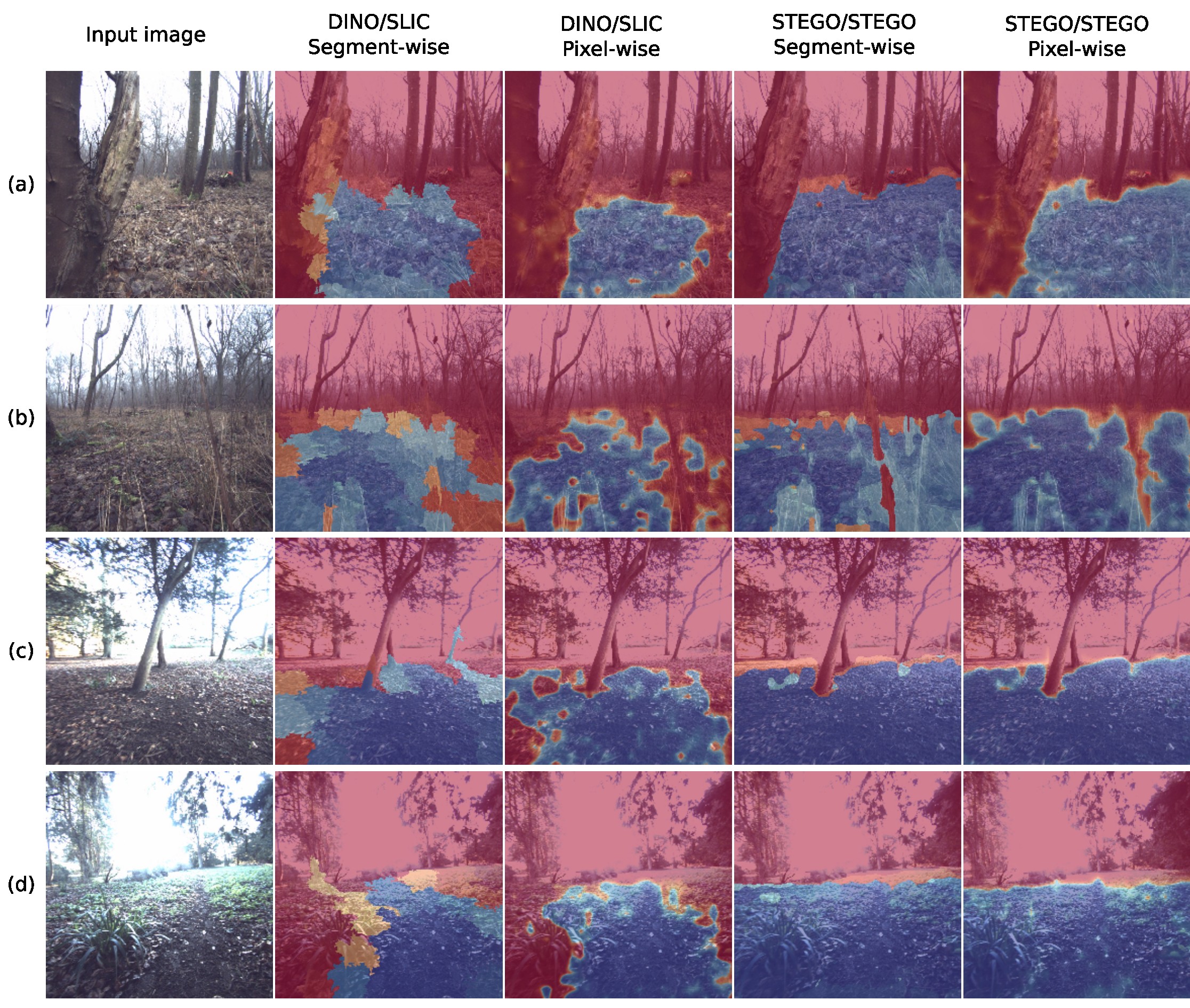}
    \caption{Inference approaches: We qualitatively compared segment-wise and pixel-wise inference using pre-trained DINO and \stego{} features. We observed advantages in executing the inference in a pixel-wise manner, which provided a fine-grained prediction regardless of the pre-trained features.
    }
    \label{fig:exp-offline-inference}
\end{figure*}

\figref{fig:exp-offline-subsampling} shows some examples of the traversability predictions when using \stego{} and SLIC segments, as well as pixel-wise segmentation. We observed consistencies between the segment-wise and pixel-wise predictions, which we explain due to the intrinsic properties of the features ( discussed in \secref{subsec:subsampling-strategies}). However, pixel-wise inference shows advantages in providing fine-grained traversability predictions, disregarding the artifacts that weak-segmentation methods such as SLIC induce, and seen in \figref{fig:exp-offline-inference} \emph{(b)} and \emph{(c)} on the tree trunk. 

\stego{} did not produce significant differences in terms of the output, consistently segmenting the traversed areas for both inference approaches. However, we did observe problems over segmenting certain areas, such as the plants in row (d), which suggest that both the features and the segments 'agreed' on the object being semantically similar to the other traversed areas.

\subsubsection{Feature Subsampling}
\label{subsec:subsampling-strategies}
\begin{figure}[t]
    \centering
    \includegraphics[width=\columnwidth]{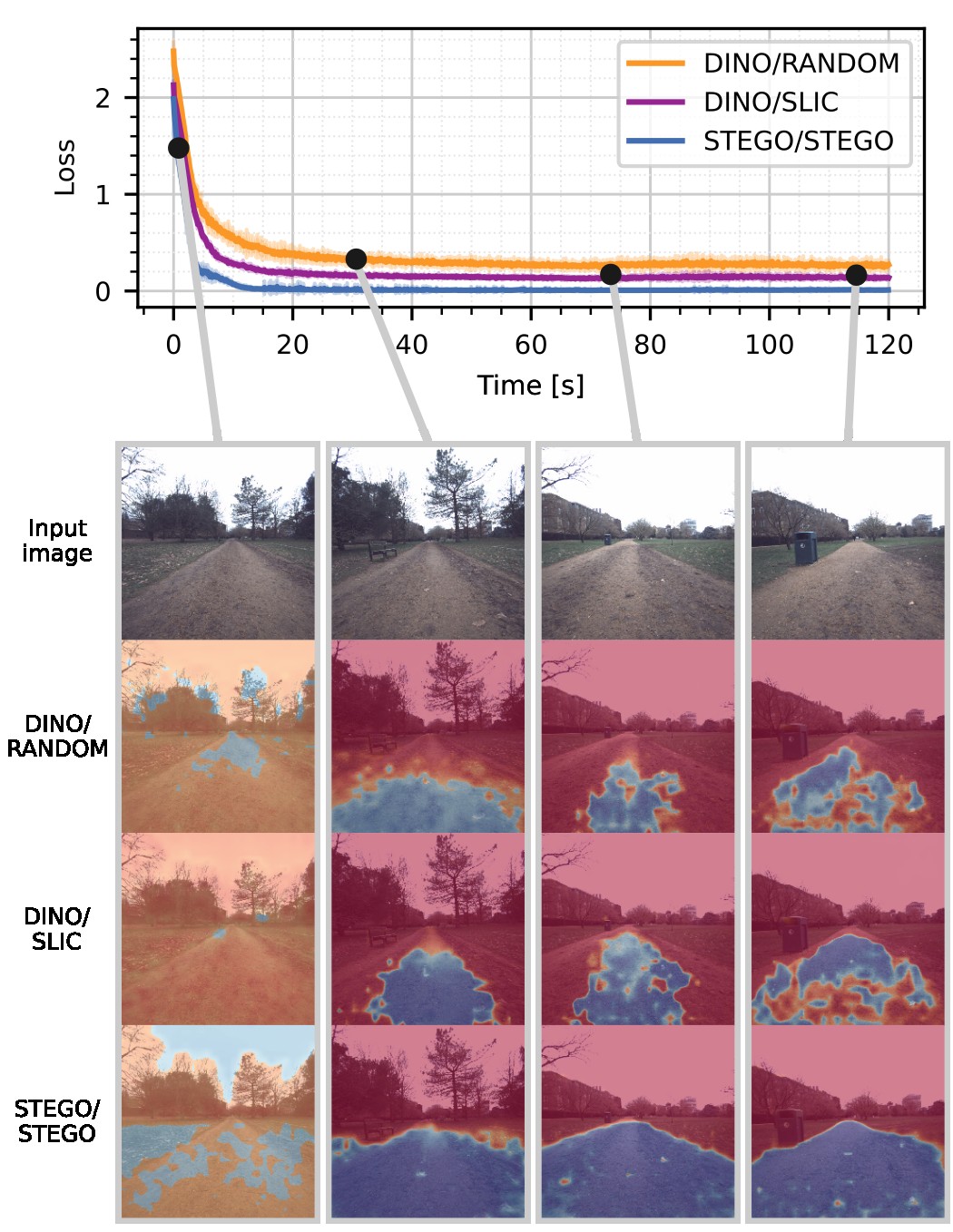}
    \caption{Feature Sub-sampling: We tested the different sub-sampling methods in the recorded path-following sequence from \secref{subsubsec:exp-real-world-path-following}. We observed that \stego{} provides significant improvements for the path-following task in both traversability prediction fidelity and training stability.
    }
    \label{fig:exp-offline-subsampling}
\end{figure}

This second experiment compared different sub-sampling strategies presented in \secref{subsubsec:feature-subsampling} in terms of traversability prediction and training. Our methodology involved re-running \gls{wvn} in post-processing using the recorded signals from the \emph{Autonomous Navigation in the Park} sequence (\secref{subsubsec:exp-real-world-path-following}). We executed five runs for each case, training the traversability prediction model from scratch without any pre-trained weights. We recorded the produced traversability predictions as well as the learning curves. \figref{fig:exp-offline-subsampling} shows the training loss, averaged over the five runs with $2\sigma$ confidence bounds, and qualitative examples of the traversability predictions when using the pixel-wise inference method.

We observed the most benefit when using the \stego{} segments and features, which enabled rapid adaptation in terms of segmenting the footpath as traversable (\figref{fig:exp-offline-subsampling}). This was also reflected in the corresponding loss curves, which achieved faster convergence and lower training loss than the other methods.

Regarding the two other sub-sampling methods, random and SLIC, we did not observe significant differences in the predicted traversability. This can be explained by the use of the same \dino{} features, which suggests that most of the expressive power is already encoded in the features, and the contribution of the sampling and mean averaging does not considerably affect the predictions. The main difference is the slightly improved training stability reflected on the lower confidence bounds of SLIC compared to random sub-sampling.

\section{Conclusion}
\label{sec:Conclusion}
We presented \glsfirst{wvn}, a system that leverages the latest advances in pre-trained self-supervised networks with a scheme to generate supervision signals while a robot operates, to achieve online, onboard visual traversability estimation. The fast adaptation capabilities of our system allowed us to easily deploy robots for navigation tasks in new environments after just a few minutes of learning from human demonstrations. 

We demonstrated \gls{wvn} through different real-world experiments and offline analyses, illustrating its fast adaptation capabilities, the consistency of its traversability prediction for local planning, and closed-loop navigation experiments, in both indoor and natural scenes. The experiments validated the key idea behind our approach of exploiting the semantic priors from pre-trained models, enabling fast generalization and adaptation in unseen scenarios from small data collected during demonstrations \emph{in the wild}. 

Regarding the limitations, the use of traction as the traversability score metric, as well as the closed-loop integration with the local terrain map via raycasting are the main aspects to investigate. These are some of the main open scientific and engineering questions for \gls{wvn}.

Lastly, to foster further research in the field, we provide the community with the codebase and pre-trained models for different environments as baselines.

\section*{Acknowledgments}
\label{sec:acknowledgments}
This work was supported by the Swiss National Science Foundation (SNSF) through project 188596, the National Centre of Competence in Research Robotics (NCCR Robotics), the European Union's Horizon 2020 research and innovation program under grant agreement No 101016970, No 101070405, and No 852044, and an ETH Zurich Research Grant. Jonas Frey is supported by the Max Planck ETH Center for Learning Systems. Matias Mattamala is supported by the National Agency for Research and Development (ANID) / DOCTORADO BECAS CHILE/2019 - 72200291 and NCCR Robotics. Maurice Fallon is supported by a Royal Society University Research Fellowship.

The authors also thank Benoit Casseau for technical support, and P\'ia Cort\'es-Zuleta for critically proofreading the manuscript.

\bibliography{bibliography}


\begin{thebibliography}{44}
\ifx \bisbn   \undefined \def \bisbn  #1{ISBN #1}\fi
\ifx \binits  \undefined \def \binits#1{#1}\fi
\ifx \bauthor  \undefined \def \bauthor#1{#1}\fi
\ifx \batitle  \undefined \def \batitle#1{#1}\fi
\ifx \bjtitle  \undefined \def \bjtitle#1{#1}\fi
\ifx \bvolume  \undefined \def \bvolume#1{\textbf{#1}}\fi
\ifx \byear  \undefined \def \byear#1{#1}\fi
\ifx \bissue  \undefined \def \bissue#1{#1}\fi
\ifx \bfpage  \undefined \def \bfpage#1{#1}\fi
\ifx \blpage  \undefined \def \blpage #1{#1}\fi
\ifx \burl  \undefined \def \burl#1{\textsf{#1}}\fi
\ifx \doiurl  \undefined \def \doiurl#1{\url{https://doi.org/#1}}\fi
\ifx \betal  \undefined \def \betal{\textit{et al.}}\fi
\ifx \binstitute  \undefined \def \binstitute#1{#1}\fi
\ifx \binstitutionaled  \undefined \def \binstitutionaled#1{#1}\fi
\ifx \bctitle  \undefined \def \bctitle#1{#1}\fi
\ifx \beditor  \undefined \def \beditor#1{#1}\fi
\ifx \bpublisher  \undefined \def \bpublisher#1{#1}\fi
\ifx \bbtitle  \undefined \def \bbtitle#1{#1}\fi
\ifx \bedition  \undefined \def \bedition#1{#1}\fi
\ifx \bseriesno  \undefined \def \bseriesno#1{#1}\fi
\ifx \blocation  \undefined \def \blocation#1{#1}\fi
\ifx \bsertitle  \undefined \def \bsertitle#1{#1}\fi
\ifx \bsnm \undefined \def \bsnm#1{#1}\fi
\ifx \bsuffix \undefined \def \bsuffix#1{#1}\fi
\ifx \bparticle \undefined \def \bparticle#1{#1}\fi
\ifx \barticle \undefined \def \barticle#1{#1}\fi
\bibcommenthead
\ifx \bconfdate \undefined \def \bconfdate #1{#1}\fi
\ifx \botherref \undefined \def \botherref #1{#1}\fi
\ifx \url \undefined \def \url#1{\textsf{#1}}\fi
\ifx \bchapter \undefined \def \bchapter#1{#1}\fi
\ifx \bbook \undefined \def \bbook#1{#1}\fi
\ifx \bcomment \undefined \def \bcomment#1{#1}\fi
\ifx \oauthor \undefined \def \oauthor#1{#1}\fi
\ifx \citeauthoryear \undefined \def \citeauthoryear#1{#1}\fi
\ifx \endbibitem  \undefined \def \endbibitem {}\fi
\ifx \bconflocation  \undefined \def \bconflocation#1{#1}\fi
\ifx \arxivurl  \undefined \def \arxivurl#1{\textsf{#1}}\fi
\csname PreBibitemsHook\endcsname

\bibitem[\protect\citeauthoryear{Gibson}{1979}]{Gibson1979}
\begin{bbook}
\bauthor{\bsnm{Gibson}, \binits{J.J.}}:
\bbtitle{The Ecological Approach to Visual Perception}.
\bpublisher{Houghton Mifflin},
\blocation{Boston}
(\byear{1979})
\end{bbook}
\endbibitem

\bibitem[\protect\citeauthoryear{Moravec and Elfes}{1985}]{Moravec1985}
\begin{bchapter}
\bauthor{\bsnm{Moravec}, \binits{H.}},
\bauthor{\bsnm{Elfes}, \binits{A.}}:
\bctitle{{High Resolution Maps from Wide Angle Sonar}}.
In: \bbtitle{IEEE Int. Conf. Robot. Autom. (ICRA)},
vol. \bseriesno{2},
pp. \bfpage{116}--\blpage{121}
(\byear{1985}).
\doiurl{10.1109/ROBOT.1985.1087316}
\end{bchapter}
\endbibitem

\bibitem[\protect\citeauthoryear{Miki et~al.}{2022}]{Miki2022a}
\begin{botherref}
\oauthor{\bsnm{Miki}, \binits{T.}},
\oauthor{\bsnm{Lee}, \binits{J.}},
\oauthor{\bsnm{Hwangbo}, \binits{J.}},
\oauthor{\bsnm{Wellhausen}, \binits{L.}},
\oauthor{\bsnm{Koltun}, \binits{V.}},
\oauthor{\bsnm{Hutter}, \binits{M.}}:
{Learning Robust Perceptive Locomotion for Quadrupedal Robots in the Wild}.
Sci. Robot.
\textbf{7}(62)
(2022)
\doiurl{10.1126/SCIROBOTICS.ABK2822}
\end{botherref}
\endbibitem

\bibitem[\protect\citeauthoryear{Maturana et~al.}{2017}]{Maturana2017}
\begin{bchapter}
\bauthor{\bsnm{Maturana}, \binits{D.}},
\bauthor{\bsnm{Chou}, \binits{P.-W.}},
\bauthor{\bsnm{Uenoyama}, \binits{M.}},
\bauthor{\bsnm{Scherer}, \binits{S.}}:
\bctitle{{Real-time Semantic Mapping for Autonomous Off-Road Navigation}}.
In: \bbtitle{Field and Service Robotics},
pp. \bfpage{335}--\blpage{350}
(\byear{2017}).
\doiurl{10.1007/978-3-319-67361-5\_22}
\end{bchapter}
\endbibitem

\bibitem[\protect\citeauthoryear{Wellhausen et~al.}{2020}]{Wellhausen2020}
\begin{barticle}
\bauthor{\bsnm{Wellhausen}, \binits{L.}},
\bauthor{\bsnm{Ranftl}, \binits{R.}},
\bauthor{\bsnm{Hutter}, \binits{M.}}:
\batitle{{Safe Robot Navigation Via Multi-Modal Anomaly Detection}}.
\bjtitle{{IEEE} Robot. Autom. Lett. (RA-L)}
(\byear{2020})
\doiurl{10.1109/LRA.2020.2967706}
\end{barticle}
\endbibitem

\bibitem[\protect\citeauthoryear{Wellhausen et~al.}{2019-04}]{Wellhausen2019}
\begin{barticle}
\bauthor{\bsnm{Wellhausen}, \binits{L.}},
\bauthor{\bsnm{Dosovitskiy}, \binits{A.}},
\bauthor{\bsnm{Ranftl}, \binits{R.}},
\bauthor{\bsnm{Walas}, \binits{K.}},
\bauthor{\bsnm{Cadena}, \binits{C.}},
\bauthor{\bsnm{Hutter}, \binits{M.}}:
\batitle{{Where Should I Walk? Predicting Terrain Properties from Images via
  Self-Supervised Learning}}.
\bjtitle{{IEEE} Robot. Autom. Lett. (RA-L)}
\bvolume{4}(\bissue{2}),
\bfpage{1509}--\blpage{1516}
(\byear{2019-04})
\doiurl{10.1109/LRA.2019.2895390}
\end{barticle}
\endbibitem

\bibitem[\protect\citeauthoryear{Gasparino et~al.}{2022}]{Gasparino2022}
\begin{barticle}
\bauthor{\bsnm{Gasparino}, \binits{M.V.}},
\bauthor{\bsnm{Sivakumar}, \binits{A.N.}},
\bauthor{\bsnm{Liu}, \binits{Y.}},
\bauthor{\bsnm{Velasquez}, \binits{A.E.B.}},
\bauthor{\bsnm{Higuti}, \binits{V.A.H.}},
\bauthor{\bsnm{Rogers}, \binits{J.}},
\bauthor{\bsnm{Tran}, \binits{H.}},
\bauthor{\bsnm{Chowdhary}, \binits{G.}}:
\batitle{{{WayFAST}: Navigation With Predictive Traversability in the Field}}.
\bjtitle{{IEEE} Robot. Autom. Lett. (RA-L)}
\bvolume{7}(\bissue{4}),
\bfpage{10651}--\blpage{10658}
(\byear{2022})
\doiurl{10.1109/LRA.2022.3193464}
\end{barticle}
\endbibitem

\bibitem[\protect\citeauthoryear{Kim et~al.}{2006}]{Kim2006}
\begin{bchapter}
\bauthor{\bsnm{Kim}, \binits{D.}},
\bauthor{\bsnm{Sun}, \binits{J.}},
\bauthor{\bsnm{Oh}, \binits{S.M.}},
\bauthor{\bsnm{Rehg}, \binits{J.M.}},
\bauthor{\bsnm{Bobick}, \binits{A.F.}}:
\bctitle{{Traversability Classification using Unsupervised On-line Visual
  Learning for Outdoor Robot Navigation}}.
In: \bbtitle{IEEE Int. Conf. Robot. Autom. (ICRA)},
pp. \bfpage{518}--\blpage{525}
(\byear{2006}).
\doiurl{10.1109/ROBOT.2006.1641763}
\end{bchapter}
\endbibitem

\bibitem[\protect\citeauthoryear{Hadsell et~al.}{2009}]{Hadsell2009}
\begin{barticle}
\bauthor{\bsnm{Hadsell}, \binits{R.}},
\bauthor{\bsnm{Sermanet}, \binits{P.}},
\bauthor{\bsnm{Ben}, \binits{J.}},
\bauthor{\bsnm{Erkan}, \binits{A.}},
\bauthor{\bsnm{Scoffier}, \binits{M.}},
\bauthor{\bsnm{Kavukcuoglu}, \binits{K.}},
\bauthor{\bsnm{Muller}, \binits{U.}},
\bauthor{\bsnm{LeCun}, \binits{Y.}}:
\batitle{{Learning Long-range Vision for Autonomous Off-road Driving}}.
\bjtitle{J. Field Robot.}
\bvolume{26}(\bissue{2}),
\bfpage{120}--\blpage{144}
(\byear{2009})
\doiurl{10.1002/ROB.20276}
\end{barticle}
\endbibitem

\bibitem[\protect\citeauthoryear{Frey et~al.}{2023}]{FreyMattamala2023}
\begin{bchapter}
\bauthor{\bsnm{Frey}, \binits{J.}},
\bauthor{\bsnm{Mattamala}, \binits{M.}},
\bauthor{\bsnm{Chebrolu}, \binits{N.}},
\bauthor{\bsnm{Cadena}, \binits{C.}},
\bauthor{\bsnm{Fallon}, \binits{M.}},
\bauthor{\bsnm{Hutter}, \binits{M.}}:
\bctitle{{Fast Traversability Estimation for Wild Visual Navigation}}.
In: \bbtitle{Robotics: Science and Systems (RSS)}
(\byear{2023}).
\doiurl{10.15607/RSS.2023.XIX.054}
\end{bchapter}
\endbibitem

\bibitem[\protect\citeauthoryear{Caron et~al.}{2021}]{Caron2021}
\begin{bchapter}
\bauthor{\bsnm{Caron}, \binits{M.}},
\bauthor{\bsnm{Touvron}, \binits{H.}},
\bauthor{\bsnm{Misra}, \binits{I.}},
\bauthor{\bsnm{J\'egou}, \binits{H.}},
\bauthor{\bsnm{Mairal}, \binits{J.}},
\bauthor{\bsnm{Bojanowski}, \binits{P.}},
\bauthor{\bsnm{Joulin}, \binits{A.}}:
\bctitle{{Emerging Properties in Self-Supervised Vision Transformers}}.
In: \bbtitle{Intl. Conf. on Computer Vision (ICCV)}
(\byear{2021}).
\doiurl{10.1109/ICCV48922.2021.00951}
\end{bchapter}
\endbibitem

\bibitem[\protect\citeauthoryear{Hamilton et~al.}{2022}]{Hamilton2022}
\begin{bchapter}
\bauthor{\bsnm{Hamilton}, \binits{M.}},
\bauthor{\bsnm{Zhang}, \binits{Z.}},
\bauthor{\bsnm{Hariharan}, \binits{B.}},
\bauthor{\bsnm{Snavely}, \binits{N.}},
\bauthor{\bsnm{Freeman}, \binits{W.T.}}:
\bctitle{Unsupervised semantic segmentation by distilling feature
  correspondences}.
In: \bbtitle{Intl. Conf. on Learning Representations (ICLR)}
(\byear{2022}).
\burl{https://openreview.net/forum?id=SaKO6z6Hl0c}
\end{bchapter}
\endbibitem

\bibitem[\protect\citeauthoryear{Achanta et~al.}{2012}]{Achanta2012}
\begin{barticle}
\bauthor{\bsnm{Achanta}, \binits{R.}},
\bauthor{\bsnm{Shaji}, \binits{A.}},
\bauthor{\bsnm{Smith}, \binits{K.}},
\bauthor{\bsnm{Lucchi}, \binits{A.}},
\bauthor{\bsnm{Fua}, \binits{P.}},
\bauthor{\bsnm{Süsstrunk}, \binits{S.}}:
\batitle{{SLIC Superpixels Compared to State-of-the-Art Superpixel Methods}}.
\bjtitle{{IEEE} Trans. Pattern Anal. Mach. Intell.}
\bvolume{34}(\bissue{11}),
\bfpage{2274}--\blpage{2282}
(\byear{2012})
\doiurl{10.1109/TPAMI.2012.120}
\end{barticle}
\endbibitem

\bibitem[\protect\citeauthoryear{Quigley et~al.}{2009}]{Quigley2009}
\begin{bchapter}
\bauthor{\bsnm{Quigley}, \binits{M.}},
\bauthor{\bsnm{Gerkey}, \binits{B.}},
\bauthor{\bsnm{Conley}, \binits{K.}},
\bauthor{\bsnm{Faust}, \binits{J.}},
\bauthor{\bsnm{Foote}, \binits{T.}},
\bauthor{\bsnm{Leibs}, \binits{J.}},
\bauthor{\bsnm{Berger}, \binits{E.}},
\bauthor{\bsnm{Wheeler}, \binits{R.}},
\bauthor{\bsnm{Ng}, \binits{A.}}:
\bctitle{{ROS: an open-source Robot Operating System}}.
In: \bbtitle{IEEE Int. Conf. Robot. Autom. (ICRA)}
(\byear{2009})
\end{bchapter}
\endbibitem

\bibitem[\protect\citeauthoryear{Chung et~al.}{2023}]{Chung2023}
\begin{botherref}
\oauthor{\bsnm{Chung}, \binits{T.H.}},
\oauthor{\bsnm{Orekhov}, \binits{V.}},
\oauthor{\bsnm{Maio}, \binits{A.}}:
{Into the Robotic Depths: Analysis and Insights from the DARPA Subterranean
  Challenge}.
Annual Review of Control, Robotics, and Autonomous Systems
\textbf{6}(1)
(2023)
\doiurl{10.1146/ANNUREV-CONTROL-062722-100728}
\end{botherref}
\endbibitem

\bibitem[\protect\citeauthoryear{Cao et~al.}{2022}]{Cao2022}
\begin{bchapter}
\bauthor{\bsnm{Cao}, \binits{C.}},
\bauthor{\bsnm{Zhu}, \binits{H.}},
\bauthor{\bsnm{Yang}, \binits{F.}},
\bauthor{\bsnm{Xia}, \binits{Y.}},
\bauthor{\bsnm{Choset}, \binits{H.}},
\bauthor{\bsnm{Oh}, \binits{J.}},
\bauthor{\bsnm{Zhang}, \binits{J.}}:
\bctitle{{Autonomous Exploration Development Environment and the Planning
  Algorithms}}.
In: \bbtitle{IEEE Int. Conf. Robot. Autom. (ICRA)},
pp. \bfpage{8921}--\blpage{8928}
(\byear{2022}).
\doiurl{10.1109/ICRA46639.2022.9812330}
\end{bchapter}
\endbibitem

\bibitem[\protect\citeauthoryear{Fan et~al.}{2021}]{Fan2021}
\begin{bchapter}
\bauthor{\bsnm{Fan}, \binits{D.D.}},
\bauthor{\bsnm{Otsu}, \binits{K.}},
\bauthor{\bsnm{Kubo}, \binits{Y.}},
\bauthor{\bsnm{Dixit}, \binits{A.}},
\bauthor{\bsnm{Burdick}, \binits{J.}},
\bauthor{\bsnm{Agha{-}Mohammadi}, \binits{A.}}:
\bctitle{{STEP: Stochastic Traversability Evaluation and Planning for Safe
  Off-road Navigation}}.
In: \bbtitle{Robotics: Science and Systems (RSS)}
(\byear{2021}).
\doiurl{10.15607/RSS.2021.XVII.021}
\end{bchapter}
\endbibitem

\bibitem[\protect\citeauthoryear{Chavez-Garcia
  et~al.}{2018}]{Chavez-Garcia2018}
\begin{barticle}
\bauthor{\bsnm{Chavez-Garcia}, \binits{R.O.}},
\bauthor{\bsnm{Guzzi}, \binits{J.}},
\bauthor{\bsnm{Gambardella}, \binits{L.M.}},
\bauthor{\bsnm{Giusti}, \binits{A.}}:
\batitle{{Learning Ground Traversability From Simulations}}.
\bjtitle{{IEEE} Robot. Autom. Lett. (RA-L)}
\bvolume{3}(\bissue{3}),
\bfpage{1695}--\blpage{1702}
(\byear{2018})
\doiurl{10.1109/LRA.2018.2801794}
\end{barticle}
\endbibitem

\bibitem[\protect\citeauthoryear{Yang et~al.}{2021}]{Yang2021}
\begin{bchapter}
\bauthor{\bsnm{Yang}, \binits{B.}},
\bauthor{\bsnm{Wellhausen}, \binits{L.}},
\bauthor{\bsnm{Miki}, \binits{T.}},
\bauthor{\bsnm{Liu}, \binits{M.}},
\bauthor{\bsnm{Hutter}, \binits{M.}}:
\bctitle{{Real-time Optimal Navigation Planning Using Learned Motion Costs}}.
In: \bbtitle{IEEE Int. Conf. Robot. Autom. (ICRA)},
pp. \bfpage{9283}--\blpage{9289}
(\byear{2021}).
\doiurl{10.1109/ICRA48506.2021.9561861}
\end{bchapter}
\endbibitem

\bibitem[\protect\citeauthoryear{Frey et~al.}{2022}]{Frey2022}
\begin{bchapter}
\bauthor{\bsnm{Frey}, \binits{J.}},
\bauthor{\bsnm{Hoeller}, \binits{D.}},
\bauthor{\bsnm{Khattak}, \binits{S.}},
\bauthor{\bsnm{Hutter}, \binits{M.}}:
\bctitle{{Locomotion Policy Guided Traversability Learning using Volumetric
  Representations of Complex Environments}}.
In: \bbtitle{IEEE/RSJ Intl. Conf. on Intelligent Robots and Systems (IROS)}
(\byear{2022}).
\doiurl{10.1109/IROS47612.2022.9982190}
\end{bchapter}
\endbibitem

\bibitem[\protect\citeauthoryear{Bradley et~al.}{2015}]{Bradley2015}
\begin{bchapter}
\bauthor{\bsnm{Bradley}, \binits{D.M.}},
\bauthor{\bsnm{Chang}, \binits{J.K.}},
\bauthor{\bsnm{Silver}, \binits{D.}},
\bauthor{\bsnm{Powers}, \binits{M.}},
\bauthor{\bsnm{Herman}, \binits{H.}},
\bauthor{\bsnm{Rander}, \binits{P.}},
\bauthor{\bsnm{Stentz}, \binits{A.}}:
\bctitle{{Scene understanding for a high-mobility walking robot}}.
In: \bbtitle{IEEE/RSJ Intl. Conf. on Intelligent Robots and Systems (IROS)},
pp. \bfpage{1144}--\blpage{1151}
(\byear{2015}).
\doiurl{10.1109/IROS.2015.7353514}
\end{bchapter}
\endbibitem

\bibitem[\protect\citeauthoryear{Schilling et~al.}{2017}]{Schilling2017}
\begin{bchapter}
\bauthor{\bsnm{Schilling}, \binits{F.}},
\bauthor{\bsnm{Chen}, \binits{X.}},
\bauthor{\bsnm{Folkesson}, \binits{J.}},
\bauthor{\bsnm{Jensfelt}, \binits{P.}}:
\bctitle{{Geometric and Visual Terrain Classification for Autonomous Mobile
  Navigation}}.
In: \bbtitle{IEEE/RSJ Intl. Conf. on Intelligent Robots and Systems (IROS)},
pp. \bfpage{2678}--\blpage{2684}
(\byear{2017}).
\doiurl{10.1109/IROS.2017.8206092}
\end{bchapter}
\endbibitem

\bibitem[\protect\citeauthoryear{Belter et~al.}{2019}]{Belter2019}
\begin{barticle}
\bauthor{\bsnm{Belter}, \binits{D.}},
\bauthor{\bsnm{Wietrzykowski}, \binits{J.}},
\bauthor{\bsnm{Skrzypczynski}, \binits{P.}}:
\batitle{Employing natural terrain semantics in motion planning for a
  multi-legged robot}.
\bjtitle{J. Intell. Robotic Syst.}
\bvolume{93}(\bissue{3-4}),
\bfpage{723}--\blpage{743}
(\byear{2019})
\doiurl{10.1007/S10846-018-0865-X}
\end{barticle}
\endbibitem

\bibitem[\protect\citeauthoryear{Shaban et~al.}{2022}]{Shaban2022}
\begin{bchapter}
\bauthor{\bsnm{Shaban}, \binits{A.}},
\bauthor{\bsnm{Meng}, \binits{X.}},
\bauthor{\bsnm{Lee}, \binits{J.}},
\bauthor{\bsnm{Boots}, \binits{B.}},
\bauthor{\bsnm{Fox}, \binits{D.}}:
\bctitle{{Semantic Terrain Classification for Off-Road Autonomous Driving}}.
In: \beditor{\bsnm{Faust}, \binits{A.}},
\beditor{\bsnm{Hsu}, \binits{D.}},
\beditor{\bsnm{Neumann}, \binits{G.}} (eds.)
\bbtitle{Conf. on Robot Learning (CoRL)}.
\bsertitle{Proceedings of Machine Learning Research},
vol. \bseriesno{164},
pp. \bfpage{619}--\blpage{629}
(\byear{2022})
\end{bchapter}
\endbibitem

\bibitem[\protect\citeauthoryear{Cai et~al.}{2022}]{Cai2022}
\begin{bchapter}
\bauthor{\bsnm{Cai}, \binits{X.}},
\bauthor{\bsnm{Everett}, \binits{M.}},
\bauthor{\bsnm{Fink}, \binits{J.}},
\bauthor{\bsnm{How}, \binits{J.P.}}:
\bctitle{Risk-aware off-road navigation via a learned speed distribution map}.
In: \bbtitle{IEEE/RSJ Intl. Conf. on Intelligent Robots and Systems (IROS)},
pp. \bfpage{2931}--\blpage{2937}
(\byear{2022}).
\doiurl{10.1109/IROS47612.2022.9982200}
\end{bchapter}
\endbibitem

\bibitem[\protect\citeauthoryear{Bajracharya et~al.}{2009}]{Bajracharya2009}
\begin{barticle}
\bauthor{\bsnm{Bajracharya}, \binits{M.}},
\bauthor{\bsnm{Howard}, \binits{A.}},
\bauthor{\bsnm{Matthies}, \binits{L.H.}},
\bauthor{\bsnm{Tang}, \binits{B.}},
\bauthor{\bsnm{Turmon}, \binits{M.}}:
\batitle{Autonomous off-road navigation with end-to-end learning for the lagr
  program}.
\bjtitle{J. Field Robot.}
\bvolume{26}(\bissue{1}),
\bfpage{3}--\blpage{25}
(\byear{2009})
\doiurl{10.1002/ROB.20269}
\end{barticle}
\endbibitem

\bibitem[\protect\citeauthoryear{Z{\"u}rn et~al.}{2021}]{Zurn2021}
\begin{barticle}
\bauthor{\bsnm{Z{\"u}rn}, \binits{J.}},
\bauthor{\bsnm{Burgard}, \binits{W.}},
\bauthor{\bsnm{Valada}, \binits{A.}}:
\batitle{Self-supervised visual terrain classification from unsupervised
  acoustic feature learning}.
\bjtitle{{IEEE} Trans. Robot.}
\bvolume{37}(\bissue{2}),
\bfpage{466}--\blpage{481}
(\byear{2021})
\doiurl{10.1109/TRO.2020.3031214}
\end{barticle}
\endbibitem

\bibitem[\protect\citeauthoryear{Kahn et~al.}{2021}]{Kahn2021}
\begin{barticle}
\bauthor{\bsnm{Kahn}, \binits{G.}},
\bauthor{\bsnm{Abbeel}, \binits{P.}},
\bauthor{\bsnm{Levine}, \binits{S.}}:
\batitle{{BADGR: An Autonomous Self-Supervised Learning-Based Navigation
  System}}.
\bjtitle{{IEEE} Robot. Autom. Lett. (RA-L)}
\bvolume{6}(\bissue{2}),
\bfpage{1312}--\blpage{1319}
(\byear{2021})
\doiurl{10.1109/LRA.2021.3057023}
\end{barticle}
\endbibitem

\bibitem[\protect\citeauthoryear{Sathyamoorthy
  et~al.}{2022}]{Sathyamoorthy2022}
\begin{bchapter}
\bauthor{\bsnm{Sathyamoorthy}, \binits{A.J.}},
\bauthor{\bsnm{Weerakoon}, \binits{K.}},
\bauthor{\bsnm{Guan}, \binits{T.}},
\bauthor{\bsnm{Liang}, \binits{J.}},
\bauthor{\bsnm{Manocha}, \binits{D.}}:
\bctitle{{TerraPN}: Unstructured terrain navigation using online
  self-supervised learning}.
In: \bbtitle{IEEE/RSJ Intl. Conf. on Intelligent Robots and Systems (IROS)},
pp. \bfpage{7197}--\blpage{7204}
(\byear{2022}).
\doiurl{10.1109/IROS47612.2022.9981942}
\end{bchapter}
\endbibitem

\bibitem[\protect\citeauthoryear{Guaman~Castro et~al.}{2023}]{GuamanCastro2023}
\begin{bchapter}
\bauthor{\bsnm{Guaman~Castro}, \binits{M.}},
\bauthor{\bsnm{Triest}, \binits{S.}},
\bauthor{\bsnm{Wang}, \binits{W.}},
\bauthor{\bsnm{Gregory}, \binits{J.M.}},
\bauthor{\bsnm{Sanchez}, \binits{F.}},
\bauthor{\bsnm{Rogers}, \binits{J.G.}},
\bauthor{\bsnm{Scherer}, \binits{S.}}:
\bctitle{How does it feel? self-supervised costmap learning for off-road
  vehicle traversability}.
In: \bbtitle{IEEE Int. Conf. Robot. Autom. (ICRA)},
pp. \bfpage{931}--\blpage{938}
(\byear{2023}).
\doiurl{10.1109/ICRA48891.2023.10160856} .
\bcomment{IEEE}
\end{bchapter}
\endbibitem

\bibitem[\protect\citeauthoryear{Jung et~al.}{2024}]{Jung2024}
\begin{bchapter}
\bauthor{\bsnm{Jung}, \binits{S.}},
\bauthor{\bsnm{Lee}, \binits{J.}},
\bauthor{\bsnm{Meng}, \binits{X.}},
\bauthor{\bsnm{Boots}, \binits{B.}},
\bauthor{\bsnm{Lambert}, \binits{A.}}:
\bctitle{{V-STRONG:} visual self-supervised traversability learning for
  off-road navigation}.
In: \bbtitle{IEEE Int. Conf. Robot. Autom. (ICRA)}
(\byear{2024})
\end{bchapter}
\endbibitem

\bibitem[\protect\citeauthoryear{Richter and Roy}{2017}]{Richter17}
\begin{bchapter}
\bauthor{\bsnm{Richter}, \binits{C.}},
\bauthor{\bsnm{Roy}, \binits{N.}}:
\bctitle{Safe visual navigation via deep learning and novelty detection}.
In: \bbtitle{Robotics: Science and Systems (RSS)},
\bconflocation{Cambridge, Massachusetts}
(\byear{2017}).
\doiurl{10.15607/RSS.2017.XIII.064}
\end{bchapter}
\endbibitem

\bibitem[\protect\citeauthoryear{Schmid et~al.}{2022}]{Schmid22}
\begin{bchapter}
\bauthor{\bsnm{Schmid}, \binits{R.}},
\bauthor{\bsnm{Atha}, \binits{D.}},
\bauthor{\bsnm{Schöller}, \binits{F.}},
\bauthor{\bsnm{Dey}, \binits{S.}},
\bauthor{\bsnm{Fakoorian}, \binits{S.}},
\bauthor{\bsnm{Otsu}, \binits{K.}},
\bauthor{\bsnm{Ridge}, \binits{B.}},
\bauthor{\bsnm{Bjelonic}, \binits{M.}},
\bauthor{\bsnm{Wellhausen}, \binits{L.}},
\bauthor{\bsnm{Hutter}, \binits{M.}},
\bauthor{\bsnm{Agha-mohammadi}, \binits{A.-a.}}:
\bctitle{Self-supervised traversability prediction by learning to reconstruct
  safe terrain}.
In: \bbtitle{IEEE/RSJ Intl. Conf. on Intelligent Robots and Systems (IROS)},
pp. \bfpage{12419}--\blpage{12425}
(\byear{2022}).
\doiurl{10.1109/IROS47612.2022.9981368}
\end{bchapter}
\endbibitem

\bibitem[\protect\citeauthoryear{Ji et~al.}{2022}]{Ji2022}
\begin{barticle}
\bauthor{\bsnm{Ji}, \binits{T.}},
\bauthor{\bsnm{Sivakumar}, \binits{A.N.}},
\bauthor{\bsnm{Chowdhary}, \binits{G.}},
\bauthor{\bsnm{Driggs-Campbell}, \binits{K.}}:
\batitle{{Proactive Anomaly Detection for Robot Navigation With Multi-Sensor
  Fusion}}.
\bjtitle{{IEEE} Robot. Autom. Lett. (RA-L)}
\bvolume{7}(\bissue{2}),
\bfpage{4975}--\blpage{4982}
(\byear{2022})
\doiurl{10.1109/LRA.2022.3153989}
\end{barticle}
\endbibitem

\bibitem[\protect\citeauthoryear{Seo et~al.}{2023}]{Seo2023}
\begin{barticle}
\bauthor{\bsnm{Seo}, \binits{J.}},
\bauthor{\bsnm{Kim}, \binits{T.}},
\bauthor{\bsnm{Kwak}, \binits{K.}},
\bauthor{\bsnm{Min}, \binits{J.}},
\bauthor{\bsnm{Shim}, \binits{I.}}:
\batitle{Scate: A scalable framework for self- supervised traversability
  estimation in unstructured environments}.
\bjtitle{IEEE Robotics and Automation Letters}
\bvolume{8}(\bissue{2}),
\bfpage{888}--\blpage{895}
(\byear{2023})
\doiurl{10.1109/LRA.2023.3234768}
\end{barticle}
\endbibitem

\bibitem[\protect\citeauthoryear{Katevenis et~al.}{1991}]{Katevenis1991}
\begin{barticle}
\bauthor{\bsnm{Katevenis}, \binits{M.}},
\bauthor{\bsnm{Sidiropoulos}, \binits{S.}},
\bauthor{\bsnm{Courcoubetis}, \binits{C.}}:
\batitle{Weighted round-robin cell multiplexing in a general-purpose {ATM}
  switch chip}.
\bjtitle{{IEEE} J. Sel. Areas Commun.}
\bvolume{9}(\bissue{8}),
\bfpage{1265}--\blpage{1279}
(\byear{1991})
\doiurl{10.1109/49.105173}
\end{barticle}
\endbibitem

\bibitem[\protect\citeauthoryear{Lee et~al.}{2017}]{Lee2017}
\begin{barticle}
\bauthor{\bsnm{Lee}, \binits{H.}},
\bauthor{\bsnm{Kwak}, \binits{K.}},
\bauthor{\bsnm{Jo}, \binits{S.}}:
\batitle{{An Incremental Nonparametric {Bayesian} Clustering-based Traversable
  Region Detection Method}}.
\bjtitle{Auton. Robot.}
\bvolume{41}(\bissue{4}),
\bfpage{795}--\blpage{810}
(\byear{2017})
\doiurl{10.1007/S10514-016-9588-7}
\end{barticle}
\endbibitem

\bibitem[\protect\citeauthoryear{Cai et~al.}{2023}]{Cai2023}
\begin{botherref}
\oauthor{\bsnm{Cai}, \binits{X.}},
\oauthor{\bsnm{Ancha}, \binits{S.}},
\oauthor{\bsnm{Sharma}, \binits{L.}},
\oauthor{\bsnm{Osteen}, \binits{P.R.}},
\oauthor{\bsnm{Bucher}, \binits{B.}},
\oauthor{\bsnm{Phillips}, \binits{S.}},
\oauthor{\bsnm{Wang}, \binits{J.}},
\oauthor{\bsnm{Everett}, \binits{M.}},
\oauthor{\bsnm{Roy}, \binits{N.}},
\oauthor{\bsnm{How}, \binits{J.P.}}:
{EVORA:} deep evidential traversability learning for risk-aware off-road
  autonomy.
CoRR
\textbf{abs/2311.06234}
(2023)
\doiurl{10.48550/ARXIV.2311.06234}
\end{botherref}
\endbibitem

\bibitem[\protect\citeauthoryear{Kingma and Ba}{2015}]{Kingma2015}
\begin{bchapter}
\bauthor{\bsnm{Kingma}, \binits{D.P.}},
\bauthor{\bsnm{Ba}, \binits{J.}}:
\bctitle{{Adam: A Method for Stochastic Optimization}}.
In: \bbtitle{Intl. Conf. on Learning Representations (ICLR)}
(\byear{2015})
\end{bchapter}
\endbibitem

\bibitem[\protect\citeauthoryear{Miki et~al.}{2022}]{Miki2022b}
\begin{bchapter}
\bauthor{\bsnm{Miki}, \binits{T.}},
\bauthor{\bsnm{Wellhausen}, \binits{L.}},
\bauthor{\bsnm{Grandia}, \binits{R.}},
\bauthor{\bsnm{Jenelten}, \binits{F.}},
\bauthor{\bsnm{Homberger}, \binits{T.}},
\bauthor{\bsnm{Hutter}, \binits{M.}}:
\bctitle{{Elevation Mapping for Locomotion and Navigation using GPU}}.
In: \bbtitle{IEEE/RSJ Intl. Conf. on Intelligent Robots and Systems (IROS)},
pp. \bfpage{2273}--\blpage{2280}
(\byear{2022}).
\doiurl{10.1109/IROS47612.2022.9981507}
\end{bchapter}
\endbibitem

\bibitem[\protect\citeauthoryear{Erni et~al.}{2023}]{Erni2023}
\begin{bchapter}
\bauthor{\bsnm{Erni}, \binits{G.}},
\bauthor{\bsnm{Frey}, \binits{J.}},
\bauthor{\bsnm{Miki}, \binits{T.}},
\bauthor{\bsnm{Mattamala}, \binits{M.}},
\bauthor{\bsnm{Hutter}, \binits{M.}}:
\bctitle{{MEM: Multi-Modal Elevation Mapping for Robotics and Learning}}.
In: \bbtitle{IEEE/RSJ Intl. Conf. on Intelligent Robots and Systems (IROS)}
(\byear{2023}).
\doiurl{10.1109/IROS55552.2023.10342108}
\end{bchapter}
\endbibitem

\bibitem[\protect\citeauthoryear{Mattamala et~al.}{2022}]{Mattamala2022}
\begin{barticle}
\bauthor{\bsnm{Mattamala}, \binits{M.}},
\bauthor{\bsnm{Chebrolu}, \binits{N.}},
\bauthor{\bsnm{Fallon}, \binits{M.}}:
\batitle{{An Efficient Locally Reactive Controller for Safe Navigation in
  Visual Teach and Repeat Missions}}.
\bjtitle{{IEEE} Robot. Autom. Lett. (RA-L)}
\bvolume{7}(\bissue{2}),
\bfpage{2353}--\blpage{2360}
(\byear{2022})
\doiurl{10.1109/LRA.2022.3143196}
\end{barticle}
\endbibitem

\bibitem[\protect\citeauthoryear{Paszke et~al.}{2019}]{Paszke2019}
\begin{bchapter}
\bauthor{\bsnm{Paszke}, \binits{A.}},
\bauthor{\bsnm{Gross}, \binits{S.}},
\bauthor{\bsnm{Massa}, \binits{F.}},
\bauthor{\bsnm{Lerer}, \binits{A.}},
\bauthor{\bsnm{Bradbury}, \binits{J.}},
\bauthor{\bsnm{Chanan}, \binits{G.}},
\bauthor{\bsnm{Killeen}, \binits{T.}},
\bauthor{\bsnm{Lin}, \binits{Z.}},
\bauthor{\bsnm{Gimelshein}, \binits{N.}},
\bauthor{\bsnm{Antiga}, \binits{L.}},
\bauthor{\bsnm{Desmaison}, \binits{A.}},
\bauthor{\bsnm{K\"{o}pf}, \binits{A.}},
\bauthor{\bsnm{Yang}, \binits{E.}},
\bauthor{\bsnm{DeVito}, \binits{Z.}},
\bauthor{\bsnm{Raison}, \binits{M.}},
\bauthor{\bsnm{Tejani}, \binits{A.}},
\bauthor{\bsnm{Chilamkurthy}, \binits{S.}},
\bauthor{\bsnm{Steiner}, \binits{B.}},
\bauthor{\bsnm{Fang}, \binits{L.}},
\bauthor{\bsnm{Bai}, \binits{J.}},
\bauthor{\bsnm{Chintala}, \binits{S.}}:
\bctitle{{PyTorch: An Imperative Style, High-Performance Deep Learning
  Library}}.
In: \bbtitle{Intl. Conf. on Neural Information Processing Systems (NeurIPS)}
(\byear{2019})
\end{bchapter}
\endbibitem

\bibitem[\protect\citeauthoryear{Wermelinger et~al.}{2016}]{Wermelinger2016}
\begin{bchapter}
\bauthor{\bsnm{Wermelinger}, \binits{M.}},
\bauthor{\bsnm{Fankhauser}, \binits{P.}},
\bauthor{\bsnm{Diethelm}, \binits{R.}},
\bauthor{\bsnm{Krüsi}, \binits{P.A.}},
\bauthor{\bsnm{Siegwart}, \binits{R.}},
\bauthor{\bsnm{Hutter}, \binits{M.}}:
\bctitle{{Navigation Planning for Legged Robots in Challenging Terrain}}.
In: \bbtitle{IEEE/RSJ Intl. Conf. on Intelligent Robots and Systems (IROS)}
(\byear{2016}).
\doiurl{10.1109/IROS.2016.7759199}
\end{bchapter}
\endbibitem

\end{thebibliography}

\vfill


\end{document}